\documentclass[11pt]{article}

\usepackage[utf8]{inputenc}
\usepackage[T1]{fontenc}
\usepackage{amsmath,amssymb,amsfonts}
\usepackage{graphicx}
\usepackage{hyperref}
\usepackage{url}
\usepackage{booktabs}
\usepackage{multirow}
\usepackage{float}
\usepackage{caption}
\usepackage{natbib}
\usepackage{xcolor}
\usepackage[margin=1in]{geometry}

\makeatletter
\g@addto@macro{\UrlBreaks}{\UrlOrds}
\makeatother

\hypersetup{
    colorlinks=true,
    linkcolor=blue!60!black,
    citecolor=blue!60!black,
    urlcolor=blue!60!black,
}

\title{Problems with Chinchilla Approach~2:\\
Systematic Biases in IsoFLOP Parabola Fits}

\author{
    Eric Czech\thanks{Open Athena AI Foundation} \and
    Zhiwei Xu\thanks{Department of Statistics, University of Michigan, Ann Arbor} \and
    Yael Elmatad\footnotemark[1] \and
    Yixin Wang\footnotemark[2] \and
    William Held\footnotemark[1]
}

\date{}

\begin{document}

\maketitle

\begin{abstract}
Chinchilla Approach~2 is among the most widely used methods for fitting neural
scaling laws. Its parabolic approximation introduces systematic biases in
compute-optimal allocation estimates, even on noise-free synthetic data. Applied
to published Llama~3 IsoFLOP data at open frontier compute scales, these biases
imply a parameter underallocation corresponding to 6.5\% of the
$3.8\times10^{25}$ FLOP training budget and \$1.4M (90\% CI:
\$412K--\$2.9M) in unnecessary compute at 50\% H100 MFU. Simulated multimodal
model misallocations show even greater opportunity costs due to higher loss
surface asymmetry. Three sources of this error are examined: IsoFLOP sampling
grid width (Taylor approximation accuracy), uncentered IsoFLOP sampling, and
loss surface asymmetry ($\alpha \neq \beta$). Chinchilla Approach~3 largely
eliminates these biases but is often regarded as less data-efficient, numerically
unstable, prone to local minima, and harder to implement. Each concern is shown
to be unfounded or addressable, especially when the partially linear structure
of the objective is exploited via Variable Projection, enabling unbiased
inference on all five loss surface parameters through a two-dimensional
optimization that is well-conditioned, analytically differentiable, and amenable
to dense, or even exhaustive, grid search. It may serve as a more convenient
replacement for Approach~2 or a more scalable alternative for adaptations of
Approach~3 to richer scaling law formulations. See
\url{https://github.com/Open-Athena/vpnls} for details and
\url{https://openathena.ai/scaling-law-analysis} for other results from this
study.
\end{abstract}

\section{Introduction}
\label{sec:introduction}

The Chinchilla paper~\citep{chinchilla} introduced three approaches to scaling
law estimation, and of these, Approach~2 has arguably seen the broadest
adoption. It has been used by leading AI labs including
DeepMind~\citep{chinchilla,sovit} (its creators),
Meta~\citep{llama3,optibert,beyond_language_modeling}, DeepSeek~\citep{deepseek},
Microsoft~\citep{ehr_scaling}, Amazon~\citep{il_scaling},
Waymo~\citep{waymo_scaling}, and Arc Institute~\citep{evo}, among others (see Table~\ref{tab:published_exponents}). It
is also a workhorse method for academic scaling law
studies~\citep{dit_scaling,dlm_scaling,biosignal_scaling} and
tutorials\footnote{\url{https://github.com/karpathy/nanochat/discussions/420}}
like those from Andrej Karpathy.

The method's appeal lies in its putative stability and data efficiency relative to
nonlinear optimization over all loss surface parameters. Rather than fitting
all five parameters of the loss surface simultaneously, Approach~2 targets only
the two scaling exponents, relying on second-order Taylor approximations that
reduce each IsoFLOP curve to a simple parabola. This sacrifices recovery
of the full loss surface but makes estimation simple, and enables extraction
of what are typically viewed as the most actionable scaling properties of a model and dataset. It does this
through a sequence of straightforward quadratic and linear fits, which frequently requires far more than 5 parameters
to be estimated (at least 3 per compute budget and 2 for the final power law), but elides the need for a nonlinear optimizer.

Despite this broad adoption, the sensitivity of the method's core
approximations and its behavior on loss surfaces that are less symmetric than
the original Chinchilla form (where parameter and token scaling exponents are
roughly equal) have not, to our knowledge, been studied in detail. Here we
revisit the basics of how to fit the Chinchilla loss model with high
precision and stability. We investigate this through
1)~noise-free synthetic simulations of IsoFLOP experiments,
2)~closed-form expression of one particular Approach~2 error mode,
3)~simulations with a noise model justified by examining
residuals from 6 real IsoFLOP experiments, each filtered through an extensive
8-step quality control pipeline, and
4)~comparing allocation forecasts from Approach~2 to other methods after
fitting them on Llama~3 training runs.

We also show how extrapolation errors trace back to suboptimal IsoFLOP
experiment design, and that pathologies in these designs can be observed in
real, high-profile scaling law studies even if they are difficult to quantify
precisely. Finally, we propose an alternative fitting method that is simple,
stable, and free of these biases while building on the same computational
shortcut as Approach~2: optimizing exponential terms separately from linear
terms. We call this approach Variable Projection with Non-negative Least
Squares (VPNLS).

This investigation is also motivated by a broader landscape of
\emph{analytical} extensions to the Chinchilla loss surface. A growing body of
work adds or modifies terms in the original functional form to account for
additional training configuration choices such as data
repetition~\citep{data_constrained},
overfitting~\citep{mupt}, precision~\citep{precision_scaling},
optimizers~\citep{optimizer_scaling}, MoE
sparsity~\citep{moe_memory_scaling},
pruning~\citep{pruning_scaling}, data quality~\citep{quality_scaling}, data
mixtures~\citep{optimal_data_mixtures,atlas_multilingual},
model shape/context length~\citep{icr_scaling},
non-embedding parameters~\citep{reconciling_scaling}, and downstream task
performance~\citep{ai2_task_scaling}, to name a few. These extensions prescribe
explicit functional forms rather than inferring scaling law structure
automatically, and they build directly on the Chinchilla model as a foundation.
Similar
studies~\citep{redundancy_scaling,data_filtering_scaling,moe_scaling,subgroup_scaling}
extend individual terms in isolation (e.g.\ token scaling terms alone) or
propose modified Kaplan scaling laws~\citep{kaplan_scaling}, both of which
would be relevant for comparable Chinchilla formulations as well. A fitting
method that recovers the base surface with greater scalability and stability may
therefore offer a stronger starting point for these adaptations.

\section{Preliminaries}
\label{sec:preliminaries}

Neural scaling laws describe how model performance improves with compute. The
Chinchilla loss surface models this relationship as:
\begin{equation}
    L(N, D) = E + \frac{A}{N^\alpha} + \frac{B}{D^\beta}
    \label{eq:loss_surface}
\end{equation}
where $N$ is the number of parameters, $D$ is the number of training tokens,
$E$ is the irreducible loss, and $A, B, \alpha, \beta$ capture how quickly
performance improves with scale.

Given a compute budget $C \approx 6ND$, the optimal allocation satisfies:
\begin{align}
    N^* &\propto C^a \quad \text{where} \quad a = \frac{\beta}{\alpha + \beta}
    \label{eq:optimal_n} \\
    D^* &\propto C^b \quad \text{where} \quad b = \frac{\alpha}{\alpha + \beta}
    \label{eq:optimal_d}
\end{align}
Recovering just the exponents $a$ and $b$ from empirical training runs is
useful for a variety of purposes. Two canonical approaches for this exist:

\subsection{Approach 2: IsoFLOP Parabolic Fitting}
\label{sec:approach2}

This approach takes advantage of the fact that along a fixed-compute contour
(IsoFLOP curve), loss as a function of $\log N$ is approximately parabolic near
the optimum. The procedure has three steps:
\begin{enumerate}
    \item \textbf{Sample IsoFLOP contours:} For each compute budget $C$, train
        models at various $(N, D)$ pairs satisfying $C = 6ND$.
    \item \textbf{Fit parabolas:} For each budget, fit
        $L = p(\log N)^2 + q(\log N) + r$ and extract the minimum $N^*$.
    \item \textbf{Fit power laws:} Regress $\log N^*$ against $\log C$ to
        recover the exponent $a$ (and similarly for $D^*$, $b$).
\end{enumerate}
The appeal is simplicity: only polynomial fits, no nonlinear optimization. The
parabolic approximation comes from a Taylor expansion of the loss surface
around the optimum.

\subsection{Approach 3: Direct Surface Fitting}
\label{sec:approach3}

The alternative is to fit all five parameters $(E, A, B, \alpha, \beta)$
simultaneously by minimizing the residual sum of squares (RSS) directly:
\begin{equation}
    \min_{E, A, B, \alpha, \beta} \sum_{i} \bigl( L_i - \hat{L}(N_i, D_i) \bigr)^2
    \quad \text{where} \quad
    \hat{L}(N, D) = E + \frac{A}{N^\alpha} + \frac{B}{D^\beta}
    \label{eq:rss}
\end{equation}

At least two practical issues arise with this direct formulation. First, $E$,
$A$, and $B$ must remain positive for the surface to be physically meaningful,
requiring box constraints. Second, the prediction involves a sum of
exponentials that can span many orders of magnitude, creating numerical
optimization difficulties.

The Chinchilla paper~\citep{chinchilla} addresses both issues with an adapted
formulation. A LogSumExp (LSE) reparameterization optimizes
$(e, a, b, \alpha, \beta)$ where $E = e^e$, $A = e^a$, $B = e^b$, enforcing
positivity without explicit bounds. The predicted log-loss is computed as:
\begin{equation}
    \log \hat{L}(N, D) = \mathrm{logsumexp}\bigl(e,\; a - \alpha \log N,\;
    b - \beta \log D\bigr)
    \label{eq:lse}
\end{equation}
The objective is then a Huber loss on log-predictions rather than MSE on
predictions. Operating in log-space improves numerical conditioning, and
the Huber function provides robustness to outliers from noisy training runs.

Because we evaluate methods on simulated data with known noise models and
want maximum likelihood estimates in some configurations, we use MSE rather
than the Huber objective throughout this study. This is the maximum likelihood
estimator (MLE) under the constant, additive Gaussian noise assumed in our
simulations. When fitting real data, we remove outliers before fitting in
most cases.
Figures~\ref{fig:appendix_residual_distributions}~and~\ref{fig:appendix_residual_variance} examine residual variance
across compute budgets in six real IsoFLOP experiments and find no strong
evidence that noise varies by budget, supporting this assumption. Some more details on these differences in optimization objectives,
reparameterizations, outcome scaling, and how they relate to common
implementations are discussed in
Appendix~\ref{sec:appendix_vpnls_validation} and the exponent inference
comparison (Section~\ref{sec:method_comparison_exponent_inference}).

\section{Error Costs: Misallocation at Scale}
\label{sec:error_costs}

Scaling laws serve at least two broad purposes in practice. At smaller scales
they guide research decisions like comparing architectures, datasets, and
training strategies before committing to a full run. For production models of
small to medium size, allocation planning typically targets deliberate
overtraining (training on more data than compute-optimal) rather than
compute-optimality itself, and Approach~3 or variations on it are much more
common in that setting. The compute-optimal scaling laws analyzed in this work
are most directly relevant to the largest models in frontier families, which
are still frequently trained near compute-optimality. This means that the Approach~2 biases detailed in later sections are
applicable, and may translate directly into wasted compute.

To quantify this, we assess the cost of Approach~2 misallocations at the
compute scale of Llama~3 405B ($3.8\times10^{25}$
FLOPs)~\citep{llama3}, which remains one of the most compute-intensive open
models trained as of early
2026~\citep{epochai_open_model_compute}. We also examine three additional loss
surfaces with published Approach~3 fit statistics---the original Chinchilla
surface~\citep{chinchilla}, and two multimodal models for which scaling is
substantially more asymmetric,
SODA~\citep{audio_scaling} (interleaved audio-text) and
Sparse-NMM~\citep{nmm_scaling} (native multimodal with sparse
mixture-of-experts). These surfaces span a range of exponent asymmetry ratios
($\beta/\alpha$) from near-unity to nearly 2:1, providing a view of how misallocation grows as the loss surface departs from
nearly equal scaling exponents.

We measure misallocation in terms of \textbf{Deadweight Compute Loss} (DCL),
defined as the excess FLOPs consumed by training at the
Approach~2--inferred allocation relative to the budget that would achieve the
same loss under optimal allocation. Given a compute budget $C$ and an inferred
data allocation $D^*$, the model size is forced by the constraint
$N = C / (6D^*)$. The resulting loss $L(N, D^*)$ is suboptimal. Inverting the
optimal loss function $L_{\mathrm{opt}}(C)$ yields the smaller budget
$C_{\mathrm{eq}}$ that would reach this same loss with optimal allocation;
$\mathrm{DCL} = C - C_{\mathrm{eq}}$. This is analogous to deadweight loss in
economics, where dataset and model sizes define a supply--demand tradeoff that
maximizes a ``total surplus''---or equivalently, minimizes validation
loss---only at the compute-optimal allocation. Any departure from optimality
wastes compute in the same sense that a price distortion wastes economic
surplus.

For Llama~3, we digitize individual IsoFLOP data points\footnote{\url{https://github.com/eric-czech/llama3_isoflop_extraction}} from Figure~2 of the
original paper~\citep{llama3} following the SVG-coordinate extraction method
of Besiroglu et al.~\citep{epochai_chinchilla_replication}, then fit both
Approach~2 and two Approach~3 implementations (VPNLS and direct L-BFGS-B) to
obtain competing allocations at the target budget. For Chinchilla, SODA, and
Sparse-NMM we take a simulated approach instead, generating synthetic IsoFLOP
data from each published loss surface, fitting Approach~2 to it under controlled bias conditions (narrow sampling grid, drifting sampling
centers) detailed in later sections, and comparing the resulting extrapolations. We include a simulated
Llama~3 condition as well, which provides an implicit measure of how much of
the empirical Llama~3 error is attributable to the real IsoFLOP experiment
design versus the biases we introduce systematically. Dollar costs are computed
assuming 50\% model FLOPs utilization~\citep{beyond_chinchilla} and \$2 per
H100-hour~\citep{olmo3}.

\begin{figure}[ht]
    \centering
    \includegraphics[width=\textwidth]{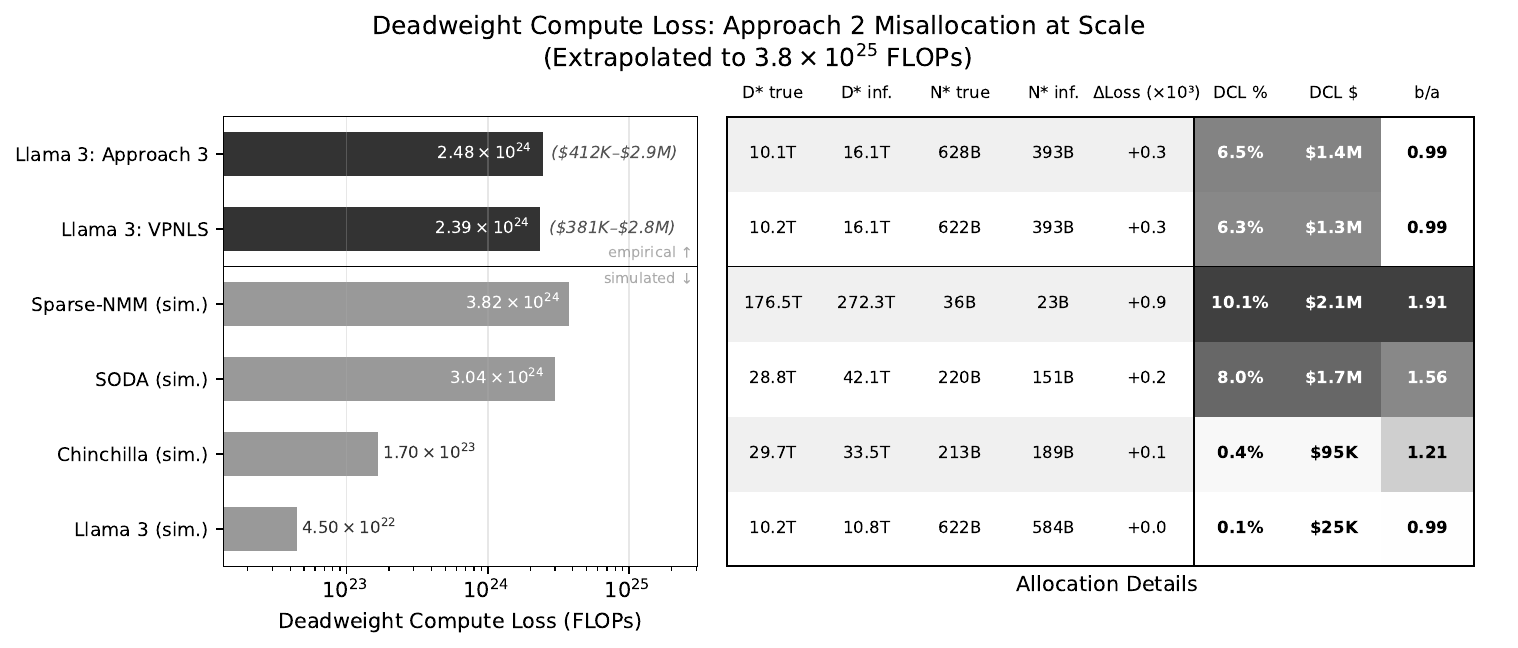}
    \caption{Approach~2 misallocation costs extrapolated to
    $3.8\times10^{25}$~FLOPs. Left: Deadweight Compute Loss (DCL) as a
    percentage of budget; dollar cost ranges for empirical rows are 90\%
    bootstrap CIs. Right: allocation details including true
    vs.\ inferred token counts and model sizes, loss penalty, and dollar cost.
    ``Empirical'' rows use Approach~2 power laws fit to digitized Llama~3
    IsoFLOP data, evaluated against VPNLS and Approach~3 surfaces.
    ``Simulated'' rows use synthetic IsoFLOP data generated from each
    published surface with the XS grid and $3\times$~drift bias used in the
    Drifting Bias simulation section (Section~\ref{sec:drifting_bias}).}
    \label{fig:dcl}
\end{figure}

Figure~\ref{fig:dcl} shows the results. On the empirical Llama~3 data,
Approach~2 overestimates the compute-optimal token count by ${\sim}60$\% relative to the Approach~3
surface, yielding DCL of 6.5\% of budget (\$1.4M).
Notably, our Approach~2 fit yields $b = 0.5368$,
which rounds exactly to $0.537$ as reported in the Llama~3 paper.
These results also include extrapolations from the Variable Projection (VPNLS)
method discussed in Section~\ref{sec:vpnls_method}. This provides a contrast
between projections made from a canonical Approach~3 implementation fit on
logarithmic loss values and one without this rescaling, a necessary assumption
for unbiased estimation in our later simulations (see Appendix~\ref{sec:appendix_vpnls_validation}).
The simulated Llama~3 condition, which isolates the systematic biases from the
IsoFLOP experiment design, produces only 0.2\% DCL (\$39K) at
$\beta/\alpha \approx 0.97$. This suggests that the magnitude of the off-center sampling bias assumed,
which matches the most extreme form of that bias used in later sections, has
little impact relative to the biases present in the actual Llama~3 IsoFLOP data
(see Figure~\ref{fig:wild}).

\begin{figure}[ht]
    \centering
    \includegraphics[width=\textwidth]{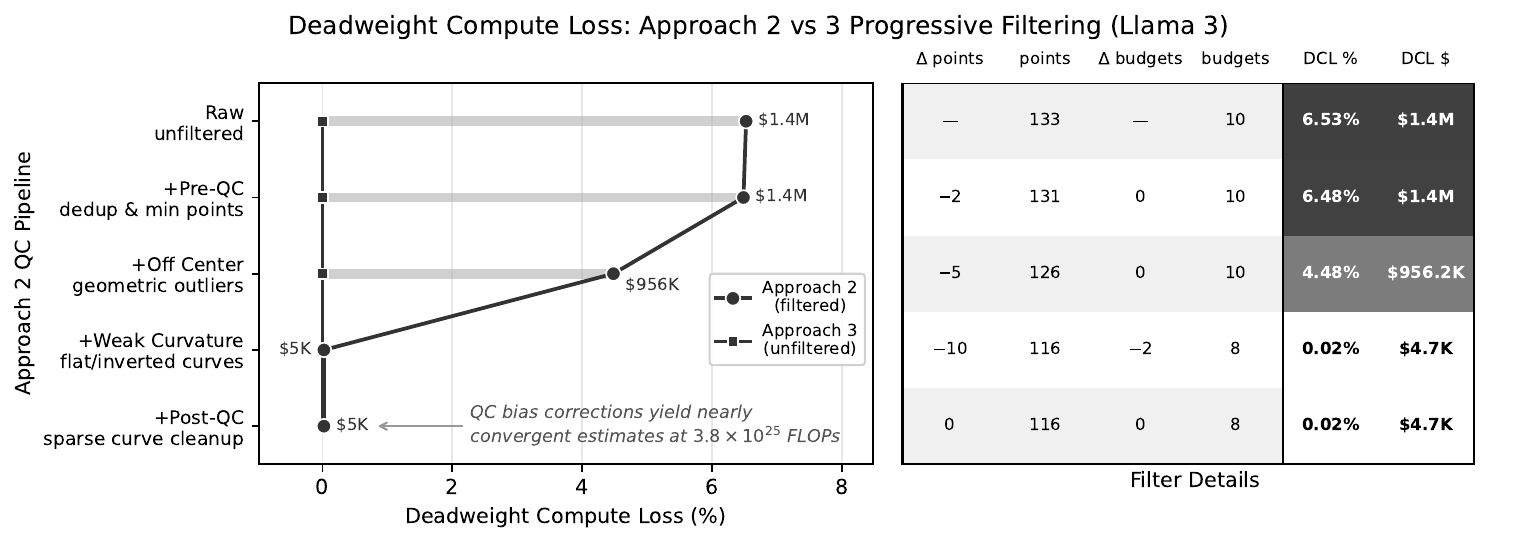}
    \caption{Effect of progressive quality control filtering on Approach~2
    Deadweight Compute Loss, measured against an Approach~3 surface fit to
    unfiltered Llama~3 data. Each row cumulatively applies one QC stage.
    Nearly all DCL reduction comes from the off-center and weak-curvature
    filters, which target specific Approach~2 biases. The right table reports points and budgets
    removed at each stage, along with DCL as a percentage of the
    $3.8\times10^{25}$~FLOP evaluation budget and dollar cost.}
    \label{fig:progressive-filter}
\end{figure}

Figure~\ref{fig:progressive-filter} shows that when known Approach~2 biases
are addressed through targeted quality control (QC), Approach~2 and Approach~3
converge on nearly identical estimates with negligible deadweight loss.
Figure~\ref{fig:appendix_isoflop_qc} visualizes that QC pipeline
across six experiments spanning different datasets and model families,
including Chinchilla, Llama~3, CoMMA, DCLM, Nemotron, and FineWeb.
Figure~\ref{fig:appendix_progressive_chinchilla} shows the same
analysis for Chinchilla data, extrapolated to its own evaluation budget of
$5.8\times10^{23}$~FLOPs. DCL drops from 37.9\% to 12.5\%, demonstrating a similar convergence
in allocation estimates.

While the simulated Llama~3 surface produces minimal waste due to its
nearly symmetric exponents, surfaces with greater asymmetry show substantially
higher costs. SODA ($\beta/\alpha \approx 1.56$) reaches
8.0\% DCL (\$1.7M) and Sparse-NMM ($\beta/\alpha \approx 1.91$) reaches
10.1\% (\$2.1M). For domains with highly asymmetric scaling exponents, which
are more common in multimodal settings~\citep{beyond_language_modeling}, Approach~2
misallocations can waste a non-trivial fraction of a large training budget.

The source of these Approach~2 misallocations is characterized in many of the
following sections. This begins with simulations of IsoFLOP experiments under
perfect, unrealistic conditions and progresses to noisier, more realistic
scenarios. Each of these demonstrates one or more estimation biases that can
compound significantly in extrapolations, as we attempted to show more directly
in this section.

\section{Error Sources: Approach 2 Biases}
\label{sec:error_sources}

\subsection{Symmetric Surfaces}
\label{sec:symmetric}

Before examining failure modes, we establish that Approach~2 works perfectly
under ideal conditions. Consider a \textbf{symmetric} loss surface where
$\alpha = \beta$:
\begin{equation}
    L(N, D) = 1.69 + \frac{400}{N^{0.31}} + \frac{400}{D^{0.31}}
    \label{eq:symmetric_surface}
\end{equation}
With equal exponents, the optimal allocation splits compute evenly between
parameters and data. The true scaling exponents are:
\begin{equation}
    a = b = \frac{0.31}{0.31 + 0.31} = 0.5
\end{equation}

We sample five IsoFLOP contours spanning $10^{17}$ to $10^{21}$ FLOPs, with 15
model sizes per curve, fit parabolas to each, and extract the optimal token
count $D^*$. Most simulations in this study use these same five compute
budgets and 15 points per IsoFLOP curve, unless stated otherwise.

\begin{figure}[ht]
    \centering
    \includegraphics[width=\textwidth]{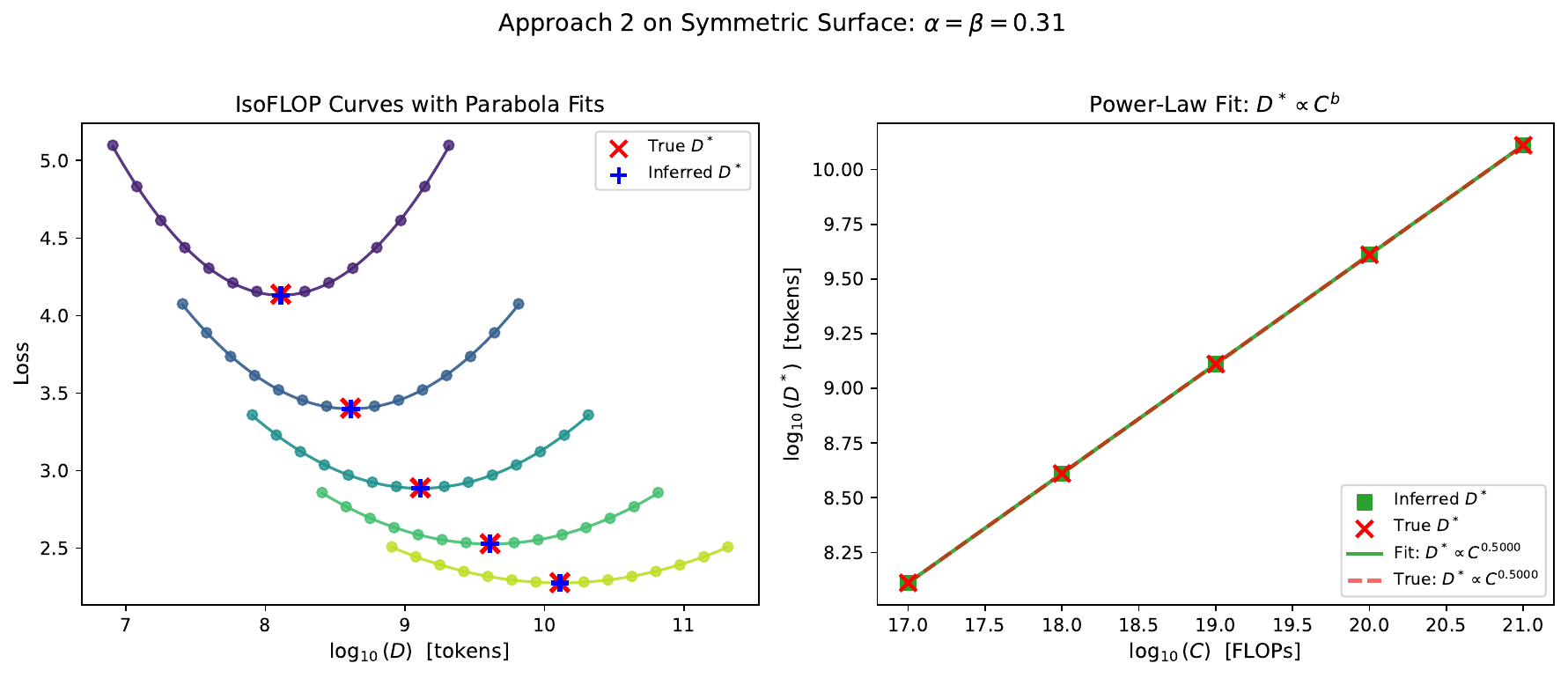}
    \caption{Approach~2 applied to a symmetric loss surface. Left: IsoFLOP
    curves with fitted parabolas. True ($\times$) and inferred ($+$) optima are
    indistinguishable. Right: Power-law fit recovers the exact scaling
    exponent.}
    \label{fig:happy_path}
\end{figure}

The results confirm perfect recovery of the token scaling exponent and
intercept as shown in Figure~\ref{fig:happy_path} and Table~\ref{tab:happy_path}. 
This establishes our baseline. Approach~2 is precisely correct under ideal
conditions that are unrealistic in practice.

\begin{table}[ht]
    \centering
    \begin{tabular}{lccc}
        \toprule
        Parameter & True Value & Inferred Value & Relative Error \\
        \midrule
        $b$ ($D^*$ exponent) & 0.500000 & 0.500000 & $+6.2 \times 10^{-12}$\% \\
        $b_0$ ($D^*$ intercept) & $-0.389076$ & $-0.389076$ & $-1.4 \times 10^{-10}$\% \\
        \bottomrule
    \end{tabular}
    \caption{Approach~2 parameter recovery on the symmetric surface.}
    \label{tab:happy_path}
\end{table}

\subsection{Asymmetric Surfaces}
\label{sec:asymmetric}

We repeat the exact same procedure as before: perfect sampling centers, no
noise, identical methodology. The only change is that the loss surface is now
\textbf{asymmetric} ($\alpha \neq \beta$).
Simulation results for this scenario show that Approach~2 produces
systematically wrong intercepts while exponents remain accurate. This means
that fitting Chinchilla Approach~2 to noise-free data drawn from the published
Chinchilla loss surface produces an incorrect scaling law, a result we found
to be counterintuitive.

We test two configurations to see how the effect scales:
\begin{itemize}
    \item \textbf{Chinchilla:} $\alpha = 0.34$, $\beta = 0.28$ (ratio $\approx 1.2$)
    \item \textbf{Asymmetric:} $\alpha = 0.46$, $\beta = 0.15$ (ratio $= 3.0$)
\end{itemize}

This ratio ($\alpha/\beta = 3$) for the Asymmetric surface is on the upper end
of previously reported statistics without being unrealistic.
Table~\ref{tab:published_exponents} reports scaling statistics for many datasets
and architectures, several of which indicate ratios in the 2--3 range (or even
higher in one case~\citep{protein_plm_scaling}). In text datasets, the asymmetry
typically runs in the opposite direction from our hypothetical Asymmetric surface
($\beta > \alpha$ rather than $\alpha > \beta$), but the opposite is frequently
true in multimodal datasets~\citep{beyond_language_modeling} and it is the magnitude of the imbalance, not its
direction, that drives most of the biases studied here.

\begin{figure}[ht]
    \centering
    \includegraphics[width=\textwidth]{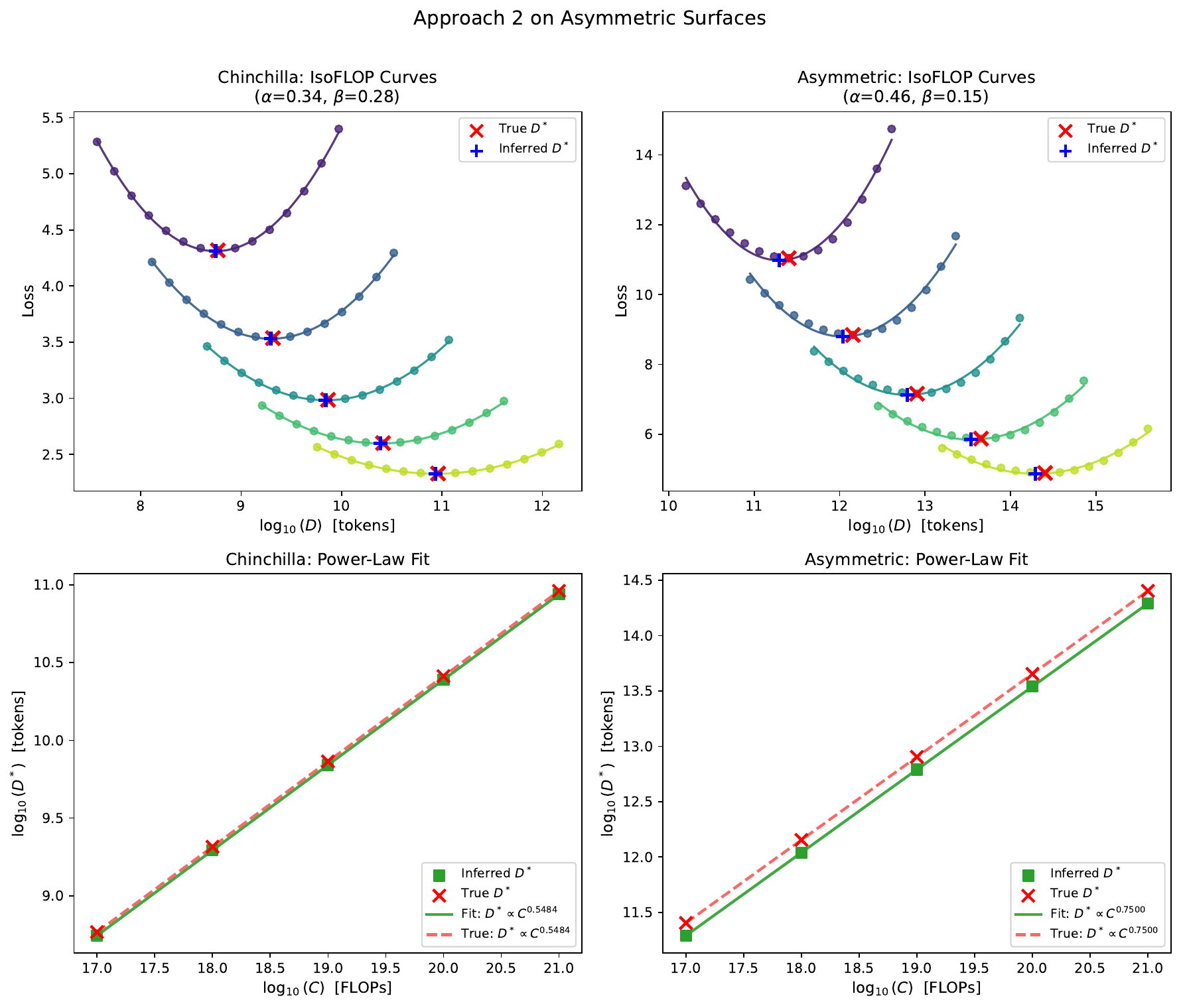}
    \caption{Approach~2 on asymmetric loss surfaces. Note the visible gap
    between true (dashed) and inferred (solid) power-law lines in the
    Asymmetric case. The exponents match perfectly, but the intercepts differ.}
    \label{fig:asymmetric}
\end{figure}

\begin{table}[H]
    \centering
    \begin{tabular}{llccc}
        \toprule
        Surface & Parameter & True Value & Inferred Value & Relative Error \\
        \midrule
        \multirow{2}{*}{Chinchilla}
        & $b$ ($D^*$ exponent) & 0.548387 & 0.548387 & $\approx 0$\% \\
        & $b_0$ ($D^*$ intercept) & $-0.555357$ & $-0.578092$ & $-4.1$\% \\
        \midrule
        \multirow{2}{*}{Asymmetric}
        & $b$ ($D^*$ exponent) & 0.750000 & 0.750000 & $\approx 0$\% \\
        & $b_0$ ($D^*$ intercept) & $-1.345791$ & $-1.459957$ & $-8.5$\% \\
        \bottomrule
    \end{tabular}
    \caption{Approach~2 parameter recovery on asymmetric surfaces.}
    \label{tab:chinchilla_surface}
\end{table}

Figure~\ref{fig:asymmetric} shows the results from this simulation, with
relative errors in Table~\ref{tab:chinchilla_surface}. Both show how errors
increase with greater asymmetry and the following section outlines how they
arise.

\subsubsection{Underlying Causes}
\label{sec:underlying_causes}

The IsoFLOP loss curve is not a true parabola.
When a parabola is fit to this curve, the parabola's minimum (vertex) does not
land exactly at the true optimum. It shifts slightly, and this shift depends
only on the loss surface shape ($\alpha$, $\beta$) and the sampling grid. It
does not depend on compute budget. The sampling grid size becomes important
here as wider grids amplify the mismatch between the true curve and its
parabolic approximation, increasing the vertex shift.

Because the IsoFLOP parabola is fit in $\log N$ space (as described in the
Approach~2 procedure), the vertex shift directly biases $N^*$. Since
$C = 6ND$, analyzing the bias in either $N^*$ or $D^*$ is sufficient, and we
focus on $N^*$ below.

Since the vertex shift is constant across all compute budgets, it biases every
inferred $N^*$ by the same multiplicative factor. When fitting $\log N^*$ vs
$\log C$ to extract scaling exponents:
\begin{itemize}
    \item The \textbf{slope (exponent)} is unchanged: multiplying all $N^*$
        values by a constant factor adds a constant to $\log N^*$, which does
        not affect the slope.
    \item The \textbf{intercept} absorbs the entire error, biased by exactly
        that multiplicative factor.
\end{itemize}

The intercept error can be derived analytically in closed form. The parabola
vertex shifts by $\delta w$ (in log-space), giving an intercept error of:
\begin{equation}
    \text{Intercept error} = 10^{\delta w} - 1
    \label{eq:intercept_error}
\end{equation}
where $\delta w = f(\alpha, \beta, W, n)$ depends only on the surface exponents
and the sampling grid (width $W$ in log-space, number of points $n$ per
IsoFLOP curve), not on $C$, $E$, $A$, or $B$. Here $W$ spans $10^{-W/2}$ to
$10^{W/2}$ times the optimal $N^*$, so $W = 2.41$ (a $\pm 16\times$ grid) means
sampling from $\frac{1}{16}\times$ to $16\times$ the optimum. Key properties:
\begin{itemize}
    \item $\delta w = 0$ when $\alpha = \beta$ (symmetric surfaces have no error)
    \item $\delta w$ grows with $|\alpha - \beta|$ (more asymmetry, more error)
    \item $\delta w$ grows with $W$ (wider sampling range, more error)
\end{itemize}

For example, with the Chinchilla parameters ($\alpha = 0.34$,
$\beta = 0.28$): a $\pm 2\times$ grid ($W = 0.60$) yields 0.3\% intercept
error, while a $\pm 16\times$ grid ($W = 2.41$) yields 4.1\% error.

The full
derivation\footnote{\url{https://github.com/Open-Athena/scaling-law-analysis/blob/main/results/article/static/scaling_parameter_errors.md}}
provides the closed-form expression for vertex shift $\delta w$ as a function
of $\alpha$, $\beta$, $W$, and $n$. It also shows how this shift translates
directly into intercept error, independent of compute budget.

\textbf{Intuition via Taylor expansion.}
A parabola is a 2nd-order polynomial, equivalent to a 2nd-order Taylor
expansion around the optimum. The approximation
$L(w) \approx L(0) + \frac{1}{2}L''(0)w^2$ is only valid when higher-order
terms are negligible, i.e., when samples are close to the true minimum. As
sampling range increases, 3rd and 4th order terms grow. For symmetric surfaces
($\alpha = \beta$), odd-order terms cancel by symmetry, preserving the vertex
location. For asymmetric surfaces, they do not cancel, shifting the fitted
vertex away from the true optimum.

\subsubsection{Error Implications}
\label{sec:error_implications}

Extrapolation to higher compute budgets requires both exponents and intercepts
to be correct. The previous subsection (Section~\ref{sec:underlying_causes})
established that asymmetric loss surfaces produce provably biased intercepts
even under ideal experimental
conditions. Here we quantify what those errors mean in practical terms by
examining compute-optimal token prediction: given a compute budget, how many
tokens does the inferred scaling law predict?

Up to this point, all analysis has assumed a single fixed sampling grid width.
We now examine how token prediction error varies with both compute budget and
sampling grid width. For surfaces with asymmetric exponents, wider sampling
grids amplify the parabola-fitting mismatch, increasing the constant vertex
shift and thus the intercept bias. To make this comparison concrete, we first
define what ``wider'' and ``narrower'' mean in quantitative terms.

A sampling grid of ``$\pm k\times$'' means the sampled values (whether model
sizes or token counts) range from $\frac{1}{k}$ to $k$ times the true optimum
at each compute budget. The total range covered is $k^2$ (the ratio of largest
to smallest), and the $\log_{10}$ of that ratio gives the number of decades the
grid spans end-to-end. Table~\ref{tab:grid_widths} shows the four grid widths
used in this analysis.

\begin{table}[ht]
    \centering
    \begin{tabular}{lcccc}
        \toprule
        Grid Name & $\pm k\times$ & Sampling Range & Total Ratio & Decade Span \\
        \midrule
        Extra Small (XS) & $\pm 2\times$ & $\frac{1}{2}\times$ to $2\times$ & $4\times$ & 0.60 \\
        Small (S) & $\pm 4\times$ & $\frac{1}{4}\times$ to $4\times$ & $16\times$ & 1.20 \\
        Large (L) & $\pm 8\times$ & $\frac{1}{8}\times$ to $8\times$ & $64\times$ & 1.81 \\
        Extra Large (XL) & $\pm 16\times$ & $\frac{1}{16}\times$ to $16\times$ & $256\times$ & 2.41 \\
        \bottomrule
    \end{tabular}
    \caption{Sampling grid widths used in simulations.}
    \label{tab:grid_widths}
\end{table}

In practice, scaling law experiments typically sample across 1 to 2 decades in
token count, placing the Small and Large grids squarely within the realistic
range. The Extra Small and Extra Large grids bracket this range on either side,
illustrating how the biases shrink or grow as the sampling window narrows or
widens. The Extra Large grid ($\pm 16\times$, $\sim$2.4 decades) is the default
used in all single-grid analyses in the preceding sections.

\begin{figure}[ht]
    \centering
    \includegraphics[width=\textwidth]{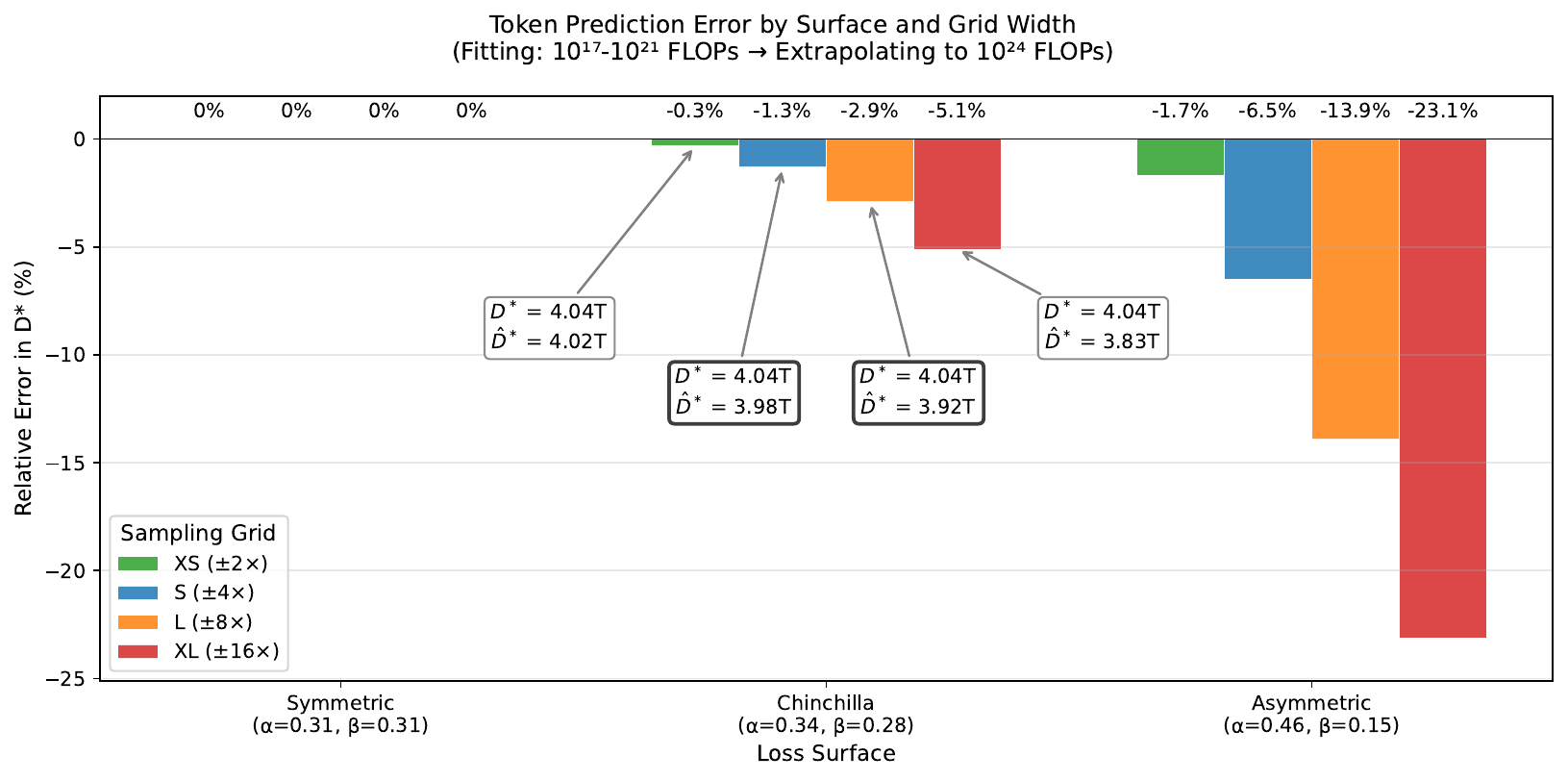}
    \caption{Relative error in compute-optimal token prediction when
    extrapolating from the training range ($10^{17}$--$10^{21}$ FLOPs) to
    $10^{24}$ FLOPs. Negative values indicate underestimation: the inferred
    scaling law predicts fewer tokens than optimal. Bars are grouped by sampling
    grid width. Annotations for the Chinchilla surface show $D^*$ (true
    compute-optimal token count) versus $\hat{D}^*$ (the Approach~2 estimate);
    the Small and Large grid annotations are emphasized (thicker borders) as they
    fall within the realistic 1--2 decade range typical of scaling law
    experiments, while Extra Small and Extra Large bracket either side as more
    extreme configurations.}
    \label{fig:extrapolation_error}
\end{figure}

The key observations from Figure~\ref{fig:extrapolation_error} are:
\begin{itemize}
    \item \textbf{Symmetric surfaces are unaffected:} When $\alpha = \beta$,
        all grid widths produce zero error.
    \item \textbf{Asymmetric surfaces underestimate:} Negative errors mean the
        inferred $D^*$ is smaller than the true $D^*$.
    \item \textbf{Wider grids amplify error:} Moving from XS ($\pm 2\times$) to
        XL ($\pm 16\times$) grids increases error from 0.3\% to 5.1\% on
        Chinchilla, and from 1.7\% to 23\% on the Asymmetric surface.
    \item \textbf{Asymmetry further amplifies error:} The Asymmetric surface
        ($\alpha/\beta = 3$) shows roughly 4--5$\times$ larger errors than
        Chinchilla at each grid width.
\end{itemize}

\subsection{Off-Center Sampling}
\label{sec:off_center}

The previous simulation sections assumed perfectly centered sampling. At every compute
budget, the IsoFLOP grid was placed exactly at the true optimum. In practice,
$N^*$ is not known before running the experiment. Sampling centers are guesses,
informed by prior estimates or heuristics, and they will likely be wrong by
some amount.

This is a distinct source of error from the asymmetry bias examined earlier.
Asymmetry errors arise from the shape of the loss surface
($\alpha \neq \beta$); off-center errors arise from where the sampling grid is
placed. To isolate this new effect, we use the symmetric surface
($\alpha = \beta = 0.31$) again where asymmetry bias is zero by construction.

\subsubsection{Constant Multiplicative Bias}
\label{sec:constant_bias}

The simplest form of off-center sampling is a constant multiplicative offset
where every compute budget's sampling center is shifted by the same factor from
the true optimum. A ``$3\times$ offset'' means each IsoFLOP grid is centered at
$3 \times D^*$ instead of $D^*$, so the grid midpoint consistently sits at
three times the true optimal token count.

Because this offset is the same at every compute budget, it has a familiar
geometric effect where each parabola vertex shifts by a constant amount in
log-space. This is the same mechanism as asymmetry bias. The slope of
$\log D^*$ vs $\log C$ is unaffected (a constant additive shift in log-space
does not change the slope), so the scaling exponent is preserved while the intercept is not.

\begin{figure}[ht]
    \centering
    \includegraphics[width=\textwidth]{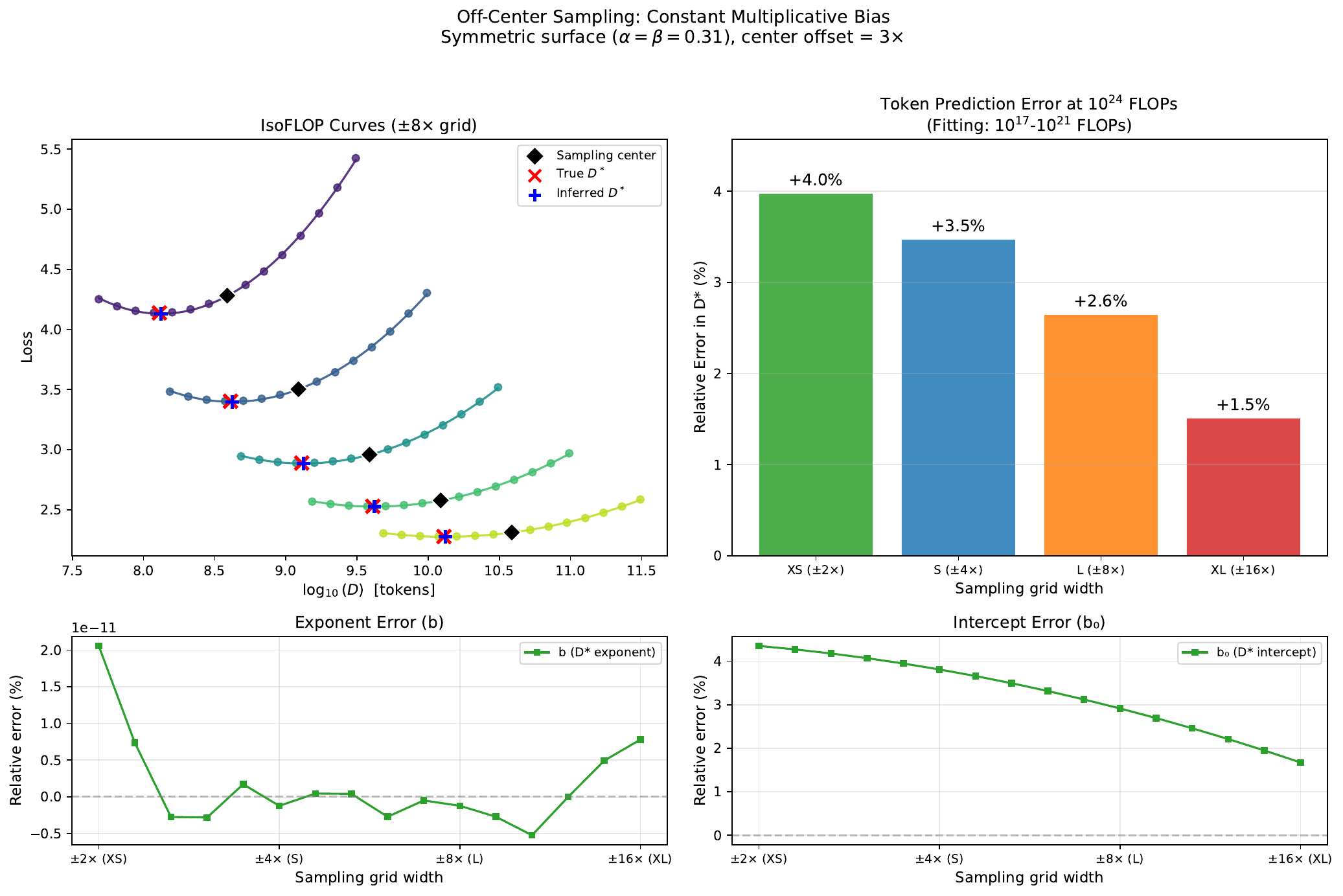}
    \caption{Effect of a constant $3\times$ offset in sampling centers on the
    symmetric surface. Top left: IsoFLOP curves at the Large grid ($\pm
    8\times$), with black diamonds marking the (off-center) sampling center, red
    $\times$ the true $D^*$, and blue $+$ the inferred $D^*$. Top right:
    extrapolation error in compute-optimal token prediction at $10^{24}$ FLOPs
    for each grid width. Bottom row: exponent and intercept errors across grid
    widths from XS ($\pm 2\times$) to XL ($\pm 16\times$), plotted on the same
    y-axis scale. The exponent is recovered perfectly (flat at zero) while the
    intercept shows systematic bias that varies with grid width.}
    \label{fig:constant_bias}
\end{figure}

The extrapolation in Figure~\ref{fig:constant_bias} (top right) shows
what this means for token prediction. All four grid widths overestimate $D^*$,
with the narrowest grid (XS) producing the largest error. This is the reverse
of the asymmetry bias pattern, where wider grids amplified error. Here, narrower
grids are more sensitive to off-center placement because fewer samples lie near
the true optimum.

The intercept error panel (bottom right) shows the same pattern across the full
continuum of grid widths. The error is always positive (the inferred $D^*$
overshoots) and decreases monotonically as the grid widens, reflecting how a
wider sampling range brings more of the true loss curve's shape into the fit,
partially compensating for the misplaced center.

\subsubsection{Drifting Bias}
\label{sec:drifting_bias}

When the offset varies with compute budget, a different failure mode occurs,
and we illustrate this by applying a multiplicative drift. The sampling center
starts at the true optimum for the lowest budget and drifts to $3\times$ the
true optimum at the highest budget, interpolating linearly in log-compute
space.

\begin{figure}[ht]
    \centering
    \includegraphics[width=\textwidth]{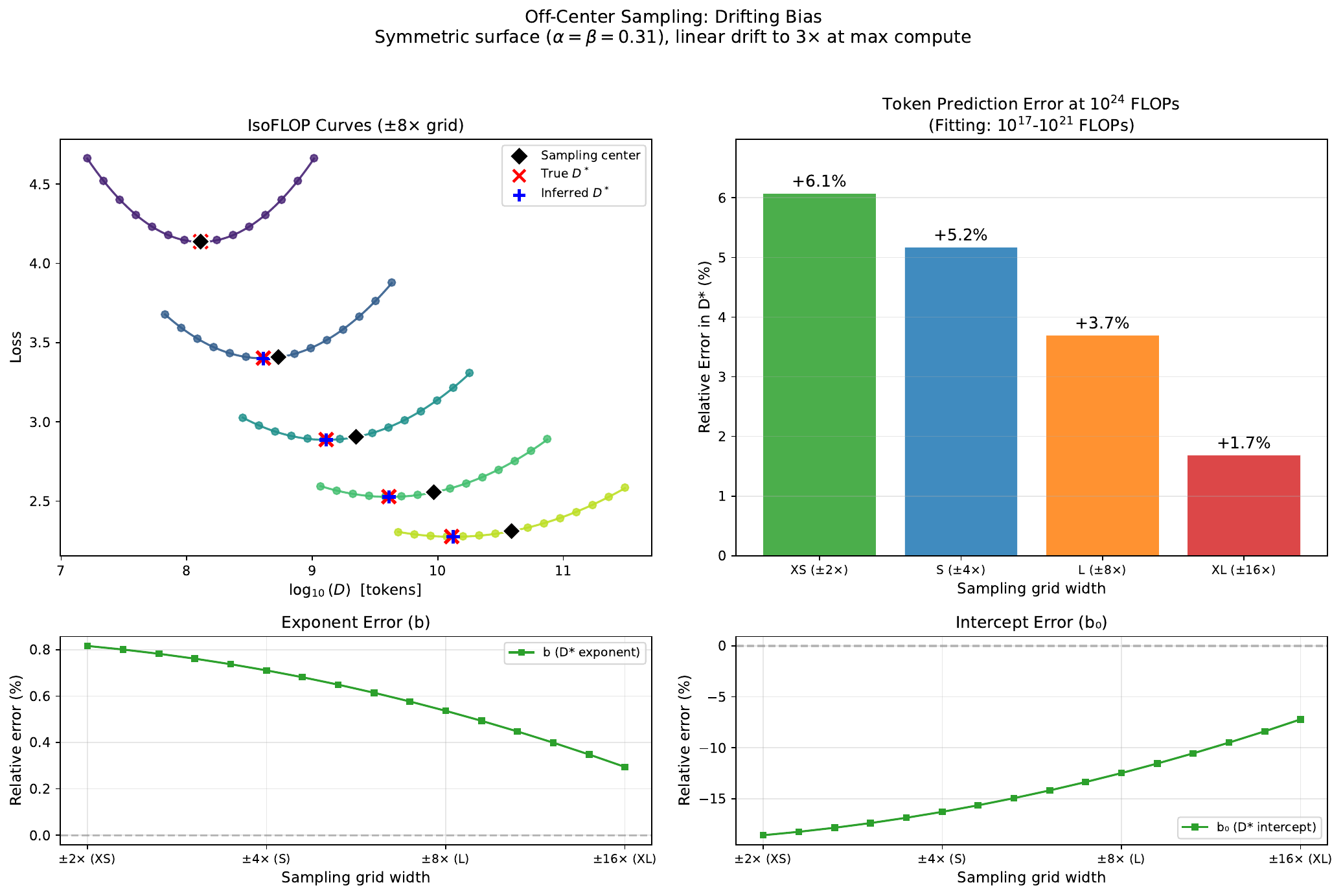}
    \caption{Effect of a linear drift in sampling centers (centered at true
    optimum for lowest budget, drifting to $3\times$ at highest budget) on the
    symmetric surface. Unlike the constant bias case, the exponent error (bottom
    left) is now non-zero: the slope of $\log D^*$ vs $\log C$ is distorted
    because the offset varies across compute budgets.}
    \label{fig:drifting_bias}
\end{figure}

Because the offset now differs across compute budgets, it no longer cancels in
the slope of $\log D^*$ vs $\log C$. Both the exponent and the intercept are
affected, as seen in Figure~\ref{fig:drifting_bias}. The bottom-left panels of Figures~\ref{fig:constant_bias}
and~\ref{fig:drifting_bias} highlight the most important difference between
compute-independent (constant bias) and compute-dependent offsets (drift bias).

\subsection{Real IsoFLOP Curves}
\label{sec:real_isoflop}

\begin{figure}[H]
    \centering
    \includegraphics[width=\textwidth]{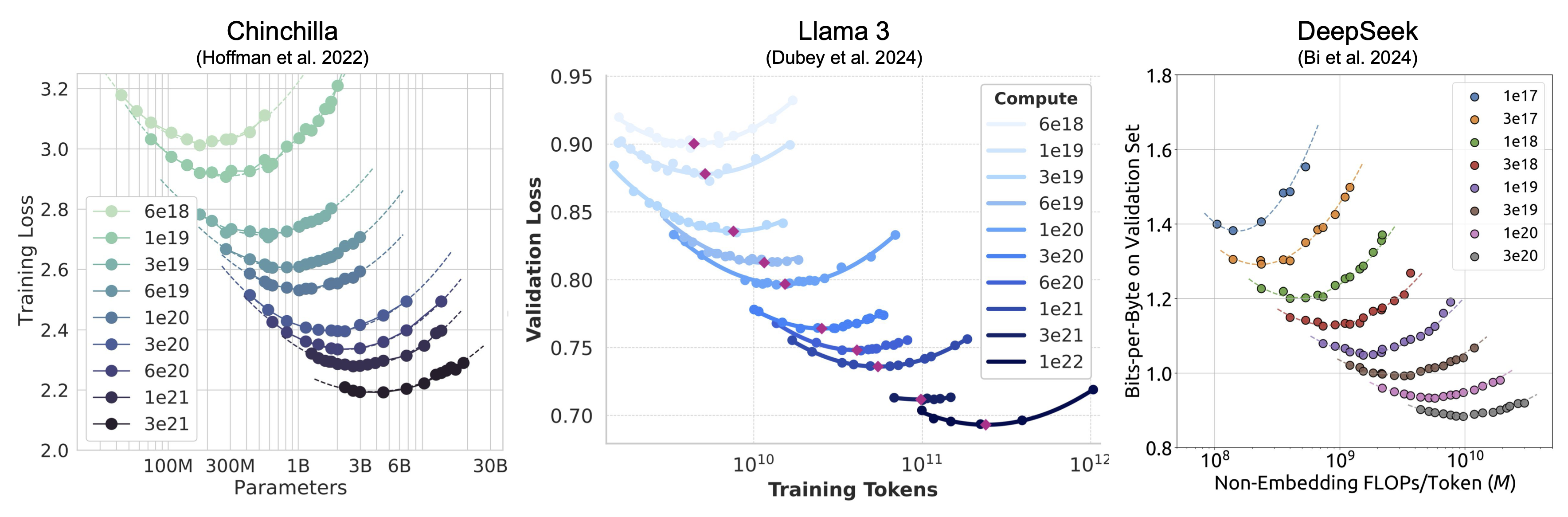}
    \caption{IsoFLOP curves from three published scaling law studies. Left:
    Chinchilla (loss vs parameters). Center: Llama~3 (loss
    vs tokens). Right: DeepSeek (bits-per-byte vs parameters). Each
    panel shows curves at multiple compute budgets, fit using Approach~2.}
    \label{fig:wild}
\end{figure}

The previous sections used synthetic, noise-free simulations to isolate
Approach~2's biases under controlled conditions. A good question is whether
the conditions that trigger these biases, asymmetric loss surfaces and
imperfectly centered sampling, actually arise in practice. To get a sense of
this, we can look at IsoFLOP curves published in three\footnote{These three and 10+ more can be found at \url{https://docs.google.com/document/d/1Ab3cYJJYu7t_99Idj9gpHv2icZmRTPBJ_brJxsgnVA4}.} prominent
scaling law studies~\citep{chinchilla,llama3,deepseek}
in Figure~\ref{fig:wild}.

We first consider biases that might arise from asymmetry. Published scaling law
studies frequently report unequal scaling exponents, as shown in
Table~\ref{tab:published_exponents}. Some degree of asymmetry is nearly
universal across modalities, and several reported exponent estimates reach the
same levels of imbalance as the most extreme configuration in our simulations.

Beyond surface asymmetry, the IsoFLOP curves in Figure~\ref{fig:wild} also
show visible signs of off-center sampling and drift:
\begin{itemize}
    \item \textbf{Off-center sampling:} At some compute budgets, the sampling
        grid does not appear centered at the curve minimum, placing more points
        on one side of the optimum than the other.
    \item \textbf{Drifting centers:} The degree of off-centering appears to
        vary across compute budgets rather than remaining constant, which is the
        drifting-bias pattern that distorts both exponents and intercepts.
\end{itemize}

To be clear, this is not a criticism of these studies. These are among the most
careful and influential scaling law analyses yet published. The point is a more
general one: the conditions under which Approach~2's biases take effect,
asymmetric surfaces and imperfect sampling centers, appear to be the norm
rather than the exception. The ideal conditions of
Section~\ref{sec:symmetric} (symmetric surface, perfectly centered grids)
occur rarely, if ever, in practice.

\subsubsection{Compounding Errors}
\label{sec:compounding}

Given evidence that both surface asymmetry and off-center sampling are present
in real studies, we can simulate what happens when these biases act
simultaneously. Using the same three loss surfaces from earlier sections, we
combine them with the $3\times$ drift and $3\times$ constant offset from the
off-center analysis. We fit Approach~2 on compute budgets from $10^{17}$ to
$10^{21}$ FLOPs and extrapolate $D^*$ predictions to $10^{24}$ FLOPs across
all four grid widths.

\begin{figure}[ht]
    \centering
    \includegraphics[width=\textwidth]{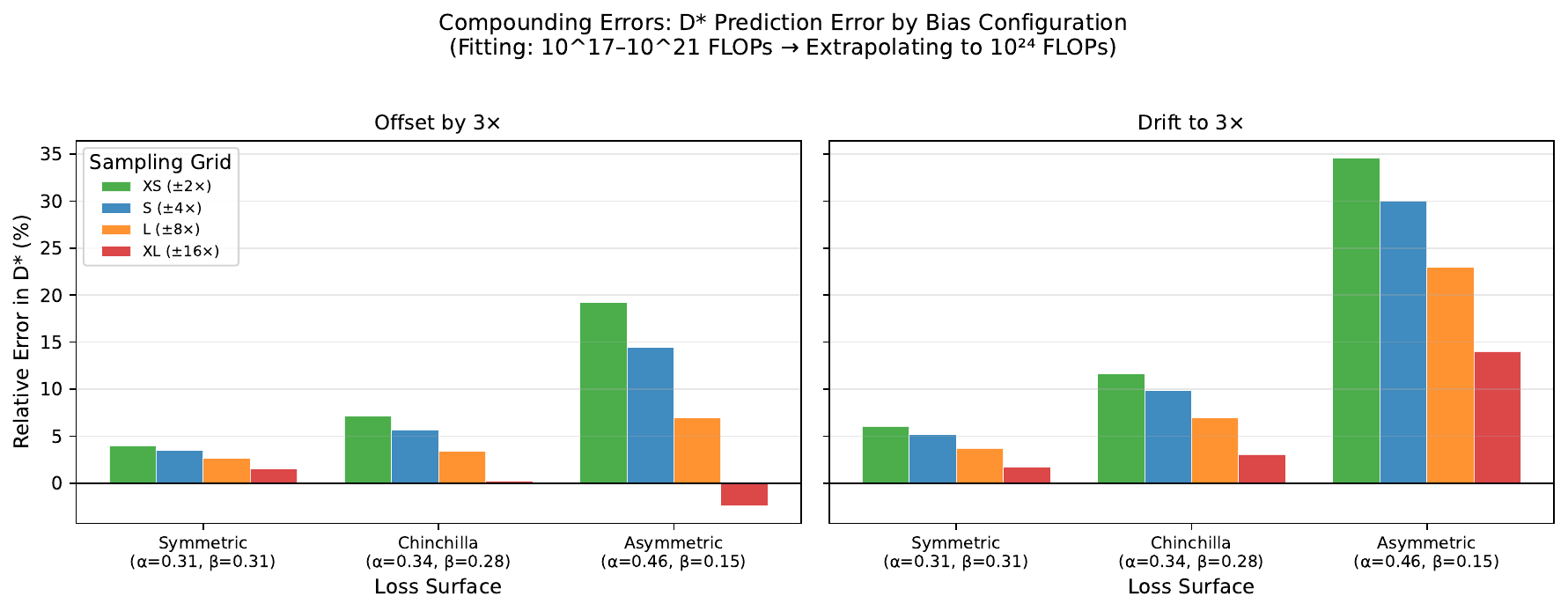}
    \caption{Relative error in $D^*$ at $10^{24}$ FLOPs with off-center
    sampling on all three loss surfaces. Left: constant $3\times$ center offset
    at every budget. Right: linear drift to $3\times$ at the highest compute
    budget. Bars are grouped by sampling grid width (XS through XL). Negative
    values indicate underestimation; positive values indicate overestimation. On
    the symmetric surface, the constant offset results correspond to
    Figure~\ref{fig:constant_bias} and the drift results correspond to
    Figure~\ref{fig:drifting_bias}; the asymmetric surfaces reveal how these
    sampling biases interact with the inherent asymmetry bias.}
    \label{fig:compounding}
\end{figure}

Results from this simulation are shown in Figure~\ref{fig:compounding}.
Compared to the baseline in Figure~\ref{fig:extrapolation_error}, where
asymmetry bias alone produces errors up to $-5$\% on Chinchilla and $-23$\% on
the Asymmetric surface, the two bias sources interact in opposite directions.
Off-center sampling pushes errors positive (overestimating $D^*$), while
asymmetry bias pushes errors negative (underestimating). When the two sources
reinforce rather than cancel, errors grow substantially---reaching nearly $+35$\% at a
maximum on the Asymmetric surface.

Figure~\ref{fig:appendix_combined_extrapolation} provides a
more detailed view of these results. It shows how $D^*$ extrapolation errors
change across compute budgets, which sources of bias grow with extrapolation
distance (drift), which remain constant (surface asymmetry and constant
offsets), and how these patterns vary across two drift rates and center offset
magnitudes.

In summary, multiple sources of bias act simultaneously in any real experiment.
Surface asymmetry and off-center sampling can each produce meaningful errors on
their own. When they happen to act in the same direction, the combined error can
compound. On our hypothetical Asymmetric surface with drift to $3\times$,
errors reach 35\% even when using the narrowest grid where the parabolic
approximation is most accurate. When they oppose, partial cancellation can
occur, but this depends on the specific combination of surface geometry, offset
magnitude, and grid width.

\section{Robust Fits: Unbiased Estimation with Linear Separation}
\label{sec:vpnls}

The previous sections showed that Approach~2's parabolic approximation
introduces systematic biases in intercepts (from asymmetry) and potentially
exponents (from off-center sampling), and that the conditions driving these
biases are visible in published scaling law studies. The natural alternative is
Approach~3, which fits all five surface parameters $(E, A, B, \alpha, \beta)$
simultaneously via nonlinear optimization. This avoids the parabolic
approximation but introduces other problems.

\subsection{Problems with Direct Surface Fitting}
\label{sec:approach3_problems}

A recent survey of over 50 scaling law papers~\citep{misfitting} documents the
landscape of fitting practices and their failure modes. The problems described within it apply to scaling laws in general, not just
Chinchilla forms like Approach~3. Over half of the papers surveyed do not fully specify
their fitting procedure (optimizer, loss function, or initialization), which
compounds reproducibility challenges, but the papers that do provide these
details suggest several common challenges.

First, the most frequently used optimizers for scaling law fits are BFGS and L-BFGS. Some
studies use SGD-family optimizers like Adam and Adagrad, though these are noted
as sometimes poorly suited for curve fitting due to limited data efficiency.
This is because methods like BFGS incorporate second-order derivatives (Hessian
matrices) directly for greater data efficiency at the expense of memory
utilization and scalability. Pure grid search offers an even more extreme
tradeoff in this direction, which the survey notes as being employed by at
least one study~\citep{data_filtering_scaling}.

Second, a limitation common to all optimizers is sensitivity to initialization, which
is primarily a consequence of ``the difficulty of optimizing over this space,
and the presence of many local minima.'' Mitigations for this include
grid search over hundreds or thousands of starting points (running the
optimizer from each and keeping the best fit), random sampling of starting
points, evaluating a coarse grid without optimization and seeding the optimizer
from the single best candidate, or initializing from previously published
parameter values.

Third, numerical stability is a significant issue that can be framed more
concretely within the context of some of the simulations discussed in prior
sections. For example, consider the Hessian of the residual sum of
squares (RSS) objective for a five-parameter fit on noise-free data from the
hypothetical Asymmetric surface ($\alpha = 0.465$,
$\beta = 0.155$), using five IsoFLOP contours from $10^{17}$ to $10^{21}$
FLOPs with 15 points per curve. Its eigenvalues quantify how sensitive the
objective is to perturbations along each parameter
direction~\citep{hessian_optimization}, and the condition number $\kappa$ (the
ratio of the largest to the smallest eigenvalue) measures how difficult the
landscape is for gradient-based methods to navigate. For this surface, the five
eigenvalues span from approximately $8 \times 10^{-6}$ to $3 \times 10^{6}$,
giving $\kappa \approx 3.5 \times 10^{11}$. The two flattest directions
(smallest eigenvalues) point almost entirely along the linear coefficients $A$
and $B$. Near the optimum, perturbing either coefficient barely changes the
RSS, making them effectively underdetermined by the data even when the data are
perfect. The steepest directions are dominated by the scaling exponents
$\alpha$ and $\beta$, and this difference in scale between exponents and linear
coefficients is likely the primary source of numerical instability for most
optimizers.

Quasi-Newton methods like L-BFGS build an approximate inverse Hessian to scale
gradient
steps across parameter directions. When eigenvalues span 12 orders of
magnitude, the gradient signal along the flat $A$/$B$ directions is negligible
compared to the steep $\alpha$/$\beta$ directions, and convergence criteria are
often satisfied by progress in the steep directions before the flat directions
are resolved. Separating the linear parameters from the nonlinear search
eliminates the ill-conditioned directions entirely. The resulting
two-dimensional landscape over $(\alpha, \beta)$ has a Hessian condition number
of $\kappa \approx 11$ in our example, a reduction by a factor of roughly
$3 \times 10^{10}$. This motivates an algorithm that exploits the partially
linear structure of the Chinchilla loss surface to search only the
well-conditioned two-dimensional subspace.

\subsection{Variable Projection (VPNLS)}
\label{sec:vpnls_method}

The Chinchilla loss surface has a partially linear structure that can be
exploited. For any fixed values of $\alpha$ and $\beta$, the remaining
parameters $(E, A, B)$ enter the model linearly and can be solved exactly via
least squares. This is the same computational shortcut that motivates
Approach~2 (optimizing exponential terms separately from linear terms), but
applied here without the parabolic approximation.

The algorithm searches over $(\alpha, \beta)$ and, at each candidate pair,
solves for $(E, A, B)$ via least squares. A coarse $k \times k$ grid search
identifies a good starting region, and a local optimizer refines it. The
linear separation is maintained throughout. The optimizer only ever navigates
the two-dimensional $(\alpha, \beta)$ surface, never the full five-parameter
space. We term this method Variable Projection with Non-negative Least Squares
(VPNLS).

In its simplest form, VPNLS uses non-negative least squares (NNLS) for the
inner solve and a gradient-free optimizer like Nelder-Mead for the outer
search:

\begin{figure}[H]
\centering
\fbox{\small
\begin{tabular}{@{\hspace{6pt}}l@{\hspace{1.5em}}l@{\hspace{6pt}}}
\textbf{function} VPNLS(data): & \\[4pt]
\hspace*{1em}\textbf{function} objective($\alpha$, $\beta$): & \\
\hspace*{2em}$\mathbf{X} \leftarrow [\mathbf{1},\; N^{-\alpha},\; D^{-\beta}]$
    & \textit{// design matrix} \\
\hspace*{2em}$(E, A, B) \leftarrow \text{NNLS}(\mathbf{X}, L)$
    & \textit{// linear solve, $E, A, B \geq 0$} \\
\hspace*{2em}\textbf{return} $\|L - \mathbf{X} \cdot [E, A, B]\|^2$ & \\[4pt]
\hspace*{1em}$(\alpha_0, \beta_0) \leftarrow \arg\min$ objective($\alpha$, $\beta$)
    & \textit{// coarse $k \times k$ grid} \\
\hspace*{1em}$(\alpha^*, \beta^*) \leftarrow \text{NelderMead}(\text{objective},\;
    \text{start}{=}(\alpha_0, \beta_0))$
    & \textit{// refine in 2D} \\
\hspace*{1em}$(E^*, A^*, B^*) \leftarrow \text{NNLS}(\mathbf{X}(\alpha^*, \beta^*), L)$
    & \textit{// recover linear params} \\[4pt]
\hspace*{1em}\textbf{return} $(E^*, A^*, B^*, \alpha^*, \beta^*)$ & \\
\end{tabular}}
\end{figure}

Switching the inner solve from NNLS to ordinary least squares (OLS) makes the
objective differentiable and enables analytical gradients. By the envelope
theorem, the gradient of RSS with respect to $(\alpha, \beta)$ depends only on
the current residuals and design matrix, not on implicit derivatives of the
optimal $(E, A, B)$. This allows L-BFGS-B to be used with exact gradients
rather than finite differences:

\begin{figure}[H]
\centering
\fbox{\small
\begin{tabular}{@{\hspace{6pt}}l@{\hspace{1.5em}}l@{\hspace{6pt}}}
\textbf{function} VPNLS(data): & \\[4pt]
\hspace*{1em}\textbf{function} objective\_and\_grad($\alpha$, $\beta$): & \\
\hspace*{2em}$\mathbf{X} \leftarrow [\mathbf{1},\; N^{-\alpha},\; D^{-\beta}]$
    & \textit{// design matrix} \\
\hspace*{2em}$(E, A, B) \leftarrow \text{OLS}(\mathbf{X}, L)$
    & \textit{// linear solve} \\
\hspace*{2em}$r \leftarrow L - \mathbf{X} \cdot [E, A, B]$
    & \textit{// residual vector} \\
\hspace*{2em}$\partial\text{RSS}/\partial\alpha \leftarrow 2A \cdot r^\top
    (\log(N) \odot N^{-\alpha})$ & \\
\hspace*{2em}$\partial\text{RSS}/\partial\beta \leftarrow 2B \cdot r^\top
    (\log(D) \odot D^{-\beta})$ & \\
\hspace*{2em}\textbf{return} $\|r\|^2$, $[\partial\text{RSS}/\partial\alpha,\;
    \partial\text{RSS}/\partial\beta]$ & \\[4pt]
\hspace*{1em}$(\alpha_0, \beta_0) \leftarrow \arg\min$ objective($\alpha$, $\beta$)
    & \textit{// coarse $k \times k$ grid} \\
\hspace*{1em}$(\alpha^*, \beta^*) \leftarrow \text{L-BFGS-B}
    (\text{objective\_and\_grad},\; \text{start}{=}(\alpha_0, \beta_0))$
    & \textit{// $\alpha, \beta > 0$} \\
\hspace*{1em}$(E^*, A^*, B^*) \leftarrow \text{OLS}
    (\mathbf{X}(\alpha^*, \beta^*), L)$
    & \textit{// recover linear params} \\[4pt]
\hspace*{1em}\textbf{return} $(E^*, A^*, B^*, \alpha^*, \beta^*)$ & \\
\end{tabular}}
\end{figure}

Because the nonlinear search is over exponential terms only, grid density
scales with the number of exponential terms rather than the total number of
parameters. A $32^2$ grid over $(\alpha, \beta)$ provides 1,024 candidates
with fine resolution, whereas the same budget in 5D gives only
$4^5 = 1{,}024$ points spread thinly across all parameters.

The implementation of this algorithm used in the remainder of this work was
validated against \texttt{ml-scalefit}~\citep{optimal_data_mixtures}. See
Appendix~\ref{sec:appendix_vpnls_validation} for details.

\subsection{Method Comparison (Parameter Recovery)}
\label{sec:method_comparison_parameter_recovery}

\begin{figure}[ht]
    \centering
    \includegraphics[width=\textwidth]{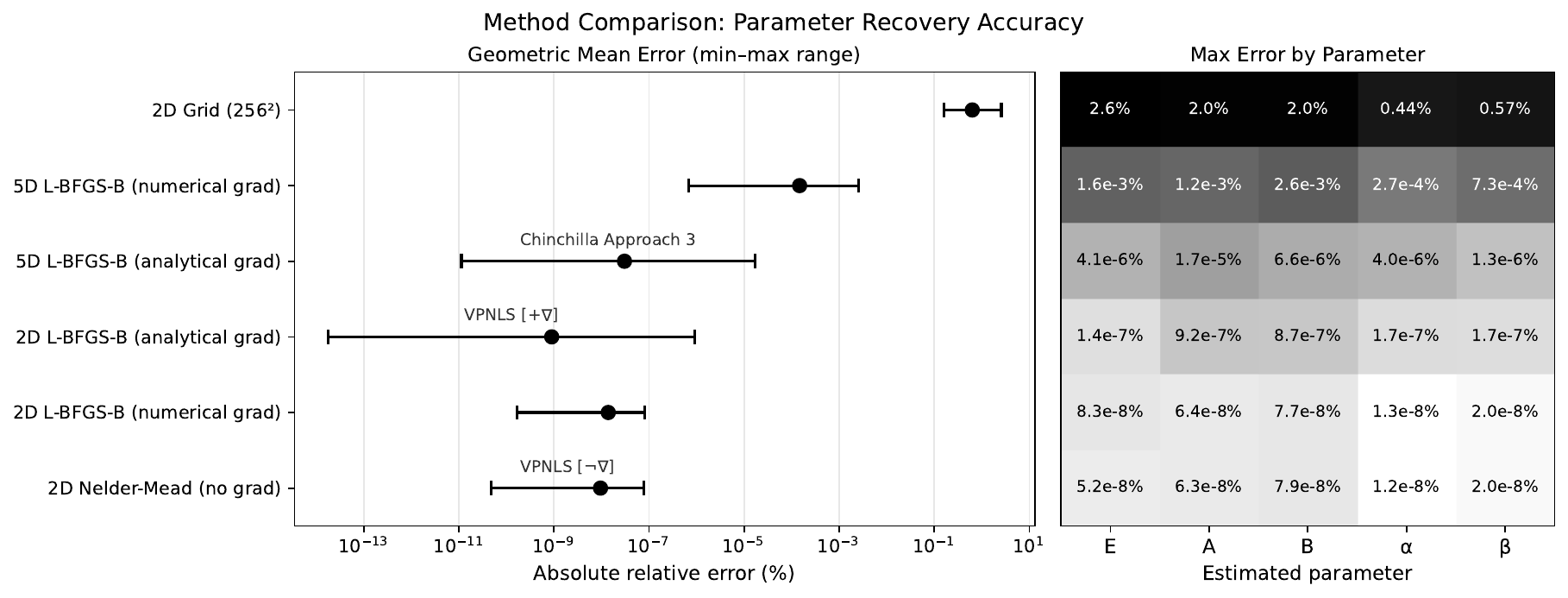}
    \caption{Comparison of six fitting methods on noise-free synthetic data
    across three loss surfaces and 20 sampling ranges (60 fits total per
    method). Left: geometric mean of $|\text{relative error}|$ (\%) pooled
    across all surfaces, grid widths, and parameters, with horizontal bars
    spanning the min-to-max range. Right: maximum $|\text{relative error}|$
    (\%) per parameter, on a log-scale colormap. Methods are sorted by
    geometric mean error, with the worst at top. VPNLS [$+\nabla$] uses
    analytical gradients (L-BFGS-B with OLS inner solve); VPNLS
    [$\lnot\nabla$] is gradient-free (Nelder-Mead with NNLS inner solve).}
    \label{fig:parameter_recovery}
\end{figure}

We compare six method configurations on noise-free synthetic data across three
loss surfaces (symmetric, Chinchilla, and high imbalance) and 20 sampling
ranges in Figure~\ref{fig:parameter_recovery}.

The configurations fall into two groups. The first uses 5D direct optimization
(Approach~3), fitting all five parameters jointly with L-BFGS-B using either
analytical or numerical (3-point central difference) gradients. The second uses
2D variable projection over $(\alpha, \beta)$ only, comparing L-BFGS-B with
analytical gradients, L-BFGS-B with numerical gradients, Nelder-Mead
(gradient-free, NNLS inner solve), and a fine $256^2$ grid search with no
local refinement. Both groups use the same total initialization budget: the 5D
methods search a $4^5 = 1{,}024$-point grid over all five parameters, while
the 2D methods search a $32^2 = 1{,}024$-point grid over $(\alpha, \beta)$
only, so that accuracy differences reflect the optimizer and loss-landscape
geometry rather than an initialization advantage.

All three locally optimized 2D variable projection methods (L-BFGS-B with
analytical gradients, L-BFGS-B with numerical gradients, and Nelder-Mead)
recover parameters to machine precision ($\sim 10^{-7}$\% or better) across
all surfaces. The well-conditioned 2D landscape makes even finite-difference
gradients reliable in the reduced search space. High-resolution grid search
($256^2$) is stable but limited by grid resolution, providing the poorest
precision among the 2D methods.

5D direct optimization (Approach~3) with analytical gradients also achieves
high precision in this noiseless setting (errors on the order of
$10^{-5}$\%), though with larger errors than any 2D method, particularly on the Asymmetric surface. The 5D numerical gradient variant performs worse, with
errors on the order of $10^{-3}$\%.
Figure~\ref{fig:appendix_method_detail} breaks this down by
surface and sampling range.

\subsection{Method Comparison (Exponent Inference)}
\label{sec:method_comparison_exponent_inference}

The parameter recovery results above are noise-free, which is obviously not
representative of practice. We now extend the compounding errors scenario
(Section~\ref{sec:compounding}; Asymmetric surface with a $3\times$ drift at
the $\pm 8\times$ grid) to a statistical setting. Gaussian noise is added to
loss values at three levels ($\sigma = 0.05, 0.1, 0.2$), applied uniformly
across all compute budgets.
Figures~\ref{fig:appendix_residual_distributions}
and~\ref{fig:appendix_residual_variance} provide empirical support for this
uniform treatment, finding no strong evidence that residual variance differs
by budget in most real IsoFLOP experiments. The number of compute
budgets varies from 2 to 4, and the number of points per IsoFLOP curve ranges
from 4 to 32. Each configuration is repeated with 256 independent noise
realizations, yielding 9{,}216 fits per method and 46{,}080 fits in total.
Figure~\ref{fig:appendix_noisy_isoflop} shows what the noisy
IsoFLOP samples look like at each noise level.

Because a scaling law study is typically run once rather than repeated hundreds
of times, sporadic optimizer failures that produce large errors in a minority
of fits are arguably the most consequential practical risk. The comparison
below emphasizes maximum errors alongside typical accuracy for this reason.

\begin{figure}[ht]
    \centering
    \includegraphics[width=\textwidth]{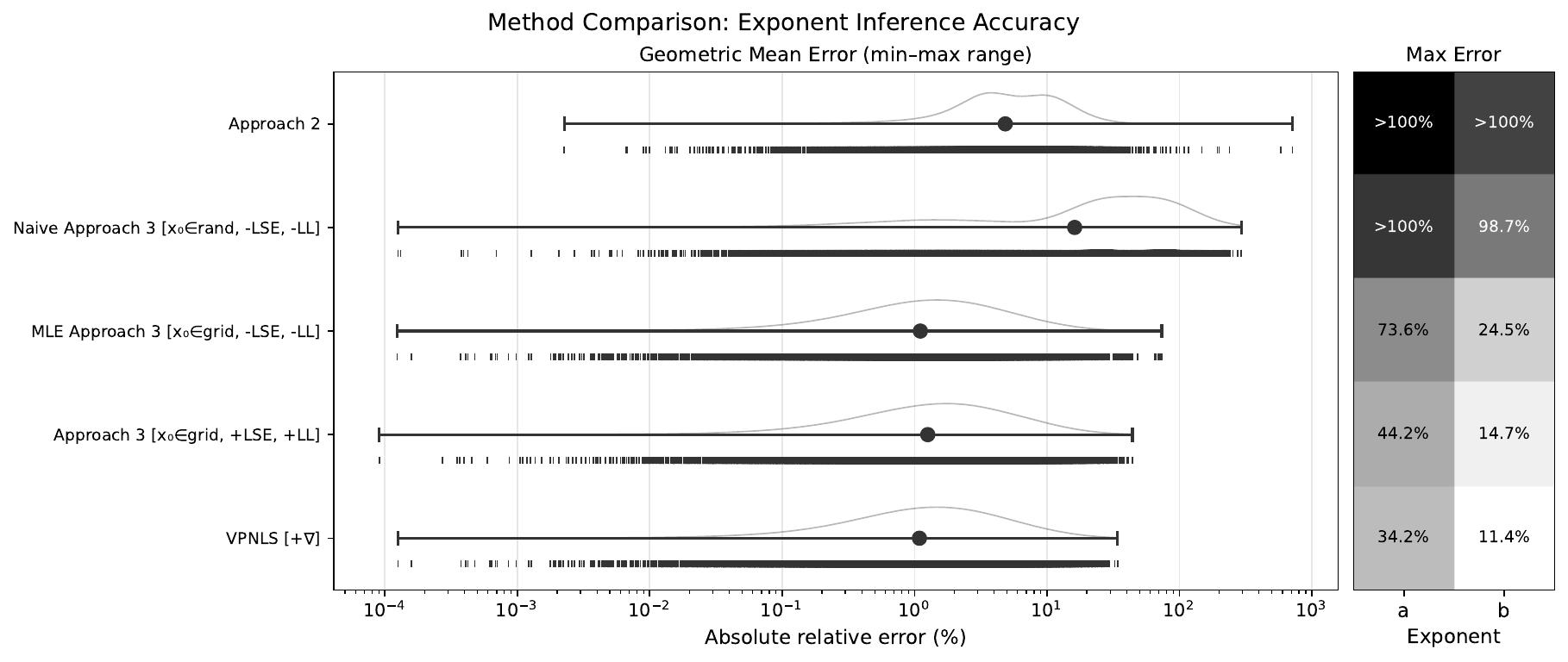}
    \caption{Comparison of five fitting methods on noisy synthetic data, pooled
    across three noise levels, three budget counts, four points-per-curve
    settings, and 256 noise realizations (9{,}216 fits per method). Left:
    geometric mean of $|\text{relative error}|$ (\%) with min-to-max range
    bars, rug ticks for individual errors, and kernel density estimate. Right:
    maximum $|\text{relative error}|$ (\%) per exponent. Methods sorted by
    worst-case error, worst at top. Surface: Asymmetric
    ($\alpha = 0.465$, $\beta = 0.155$) with $3\times$ drift, $\pm 8\times$
    grid, compute budgets $10^{17}$--$10^{21}$~FLOPs.}
    \label{fig:exponent_inference}
\end{figure}

Five methods are compared in Figure~\ref{fig:exponent_inference}. Approach~2 uses the parabolic IsoFLOP pipeline.
Naive Approach~3 uses a single random initialization without LSE
reparameterization. MLE Approach~3 uses grid initialization with the direct MSE
objective (the maximum likelihood estimator under additive Gaussian noise). The
final Approach~3 variant uses grid initialization with the LSE
reparameterization and log-loss objective, matching the formulation in the
Chinchilla paper~\citep{chinchilla}. VPNLS uses L-BFGS-B with analytical
gradients.

The parameter recovery comparison (Section~\ref{sec:method_comparison_parameter_recovery})
evaluated all five surface parameters, but Approach~2 does not estimate them
individually. Here we focus on the scaling exponents $a = \beta / (\alpha +
\beta)$ and $b = \alpha / (\alpha + \beta)$, which determine compute-optimal
allocation and are the quantities all five methods can be compared on directly.
Figure~\ref{fig:exponent_inference} pools these exponent errors across all
experimental conditions.

Approach~2 has the largest maximum errors overall (716\% on $a$, 239\% on
$b$), reflecting the structural bias from the parabolic approximation
(Section~\ref{sec:asymmetric}). Even with 32 points per curve and low noise,
the systematic inaccuracy persists. Naive Approach~3 (single random
initialization, no LSE) also produces extreme maximum errors (296\% on $a$),
confirming that without careful initialization and parameterization, 5D
optimization is unreliable.

The canonical Chinchilla Approach~3 formulation (LSE reparameterization with
log-loss objective, grid initialization) reduces maximum errors dramatically to
44\% on $a$. The LSE reparameterization enforces positivity and stabilizes
optimization. However, because the noise model in these simulations is additive
Gaussian, the log-loss objective is slightly misspecified, which is visible as a
higher geometric mean error compared to the MLE variant and VPNLS.

The MLE variant of Approach~3 (grid initialization, original-space MSE) is the
correct maximum likelihood estimator for additive Gaussian noise. In this
experiment, however, it exhibited larger maximum errors than either canonical
Approach~3 or VPNLS (74\% vs 44\% vs 34\% on the $a$ exponent), despite
slightly better typical accuracy. This is primarily due to the lack of LSE
reparameterization.

VPNLS produced the smallest maximum errors in this experiment (34\% on $a$,
11\% on $b$) with typical accuracy roughly equivalent to Approach~3. The
detailed breakdown in Figure~\ref{fig:appendix_exponent_errors} shows how
errors distribute across noise levels, budget counts, and dataset sizes.

\subsection{Method Comparison (Data Efficiency)}
\label{sec:method_comparison_data_efficiency}

\begin{figure}[ht]
    \centering
    \includegraphics[width=\textwidth]{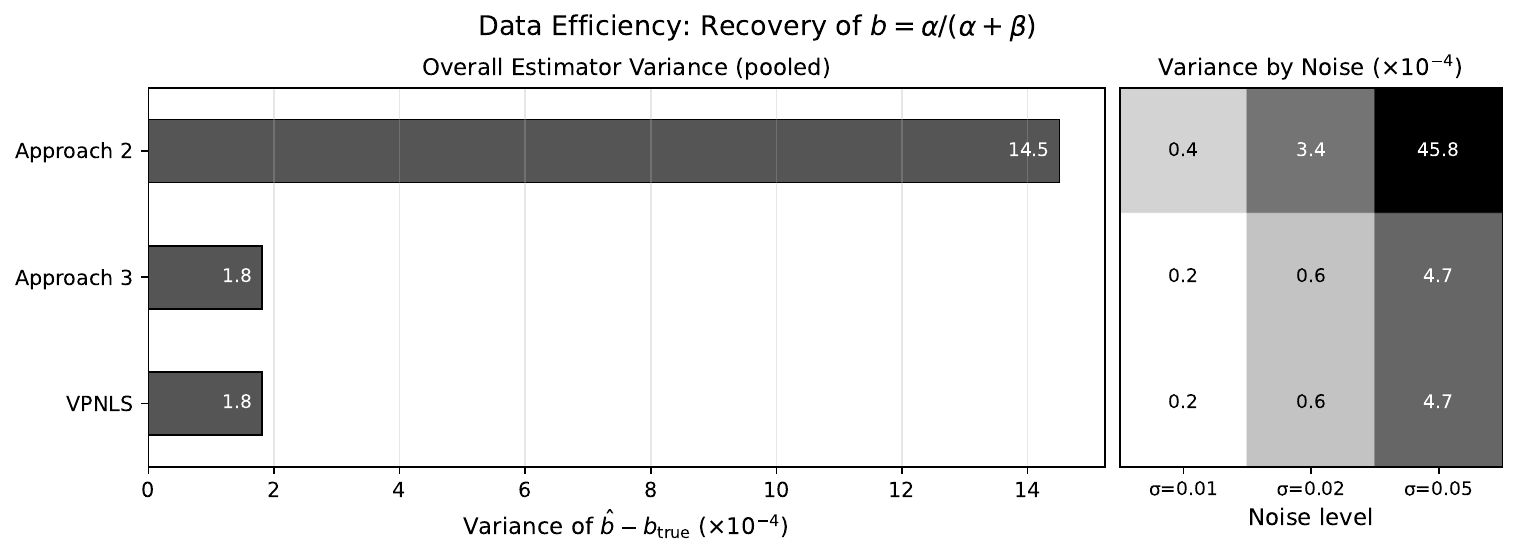}
    \caption{Estimator variance for the compute-optimal exponent
    $b = \alpha/(\alpha+\beta)$ on a symmetric surface
    ($\alpha = \beta = 0.31$, $A = B = 400$, $E = 1.69$, true $b = 0.5$) with
    perfectly centered IsoFLOP sampling. Left: overall variance pooled across
    noise levels. Right: variance by noise level. Sweep over points per curve
    $\in \{21, 31, 41\}$, budget counts $\in \{3, 5, 7\}$, log-widths
    $\in \{\pm 2\times, \pm 4\times, \pm 8\times\}$, noise levels
    $\sigma \in \{0.01, 0.02, 0.05\}$, with 10 trials per configuration.
    Lower variance is better. Approach~3 and VPNLS achieve comparable precision;
    Approach~2 is consistently less efficient.}
    \label{fig:data_efficiency}
\end{figure}

The exponent inference comparison above used an asymmetric surface with
off-center sampling, conditions designed to test structural biases. Here we do
the opposite and build on the ideal conditions of the symmetric surfaces
section ($\alpha = \beta = 0.31$, perfectly centered sampling) where Approach~2
has no structural bias. All methods are unbiased in this setting, so the
relevant comparison is variance. For unbiased estimators, statistical efficiency
is proportional to the inverse of variance, so lower variance means more
precise estimates for a given sample size. Noise levels
($\sigma \in \{0.01, 0.02, 0.05\}$) are calibrated to the empirical residual
standard deviations observed across real IsoFLOP experiments
(Figures~\ref{fig:appendix_residual_distributions}
and~\ref{fig:appendix_residual_variance}). Even under these ideal conditions,
Figure~\ref{fig:data_efficiency} shows that Approach~2 recovers scaling
exponents with roughly $8\times$ higher variance than Approach~3 or VPNLS.

Figure~\ref{fig:appendix_data_efficiency} breaks these results down by noise
level, showing the full distribution of signed errors for each method.

\section{Conclusion}
\label{sec:conclusion}

The Approach~2 biases documented in this paper are structural, not statistical.
They exist on noise-free data with perfect experimental conditions and, as the
noisy method comparison (Section~\ref{sec:method_comparison_exponent_inference})
shows, persist under realistic noise levels with varying amounts of data.

At least three sources of Approach~2 error can compound in practice. IsoFLOP
sampling grid width, uncentered IsoFLOP sampling, and loss surface asymmetry
all bias inference and extrapolations in different ways. Published IsoFLOP
curves from prominent scaling law studies
(Section~\ref{sec:real_isoflop}) show clear signs of both asymmetry and
off-center sampling. At frontier compute scales
(Section~\ref{sec:error_costs}), we estimate that these biases translate to a
potential 6.5\% decrease in training FLOPs (\$1.4M) on Llama~3 data
and potentially more on multimodal surfaces with greater asymmetry.

Approach~3 eliminates these biases. With grid initialization and LSE
reparameterization (as specified in the original Chinchilla
paper~\citep{chinchilla}), it achieves typical accuracy comparable to VPNLS in
our experiments. A recent survey~\citep{misfitting} suggests that these
important optimization details may be omitted or not reported in some studies.

VPNLS is at least as stable and accurate as more typical Approach~3
implementations while being structurally simpler. Variable projection separates
exponential from linear terms, reducing the nonlinear search to the exponential
terms only. This makes dense grid search practical because the grid grows with
the number of exponential terms rather than all parameters, and these exponents
occupy tight ranges (typically 0 to 1) unlike the linear coefficients which
span several orders of magnitude.

This structural advantage extends naturally to richer loss surface
specifications. The analytical extensions discussed in the introduction
(Section~\ref{sec:introduction}), such as epochs, data quality, and MoE
sparsity, often add linear terms that could be omitted from direct
optimization. This structural simplification may provide a more scalable
foundation for future work. It also lends well to simple implementations of
the Chinchilla loss model alone, e.g.\ it is possible to express a simplified
version of the algorithm\footnote{\url{https://gist.github.com/eric-czech/ebd9a80d58c7b5e9c40ba390ff884617}}
that guarantees globally optimal solutions at a desired level of scaling
exponent precision, uses only ${\sim}70$ lines of JavaScript code, and
requires no dependencies (no optimization or linear algebra libraries).

Lastly, practitioners using Approach~2 should be aware that scaling exponent
estimates carry a systematic bias that often grows with loss surface asymmetry,
sampling center offsets, and sampling grid width. When fitting surfaces
directly with Approach~3 instead, grid initialization and LSE
reparameterization should be used. And VPNLS offers equivalent accuracy with a
simpler optimization structure to aid in scaling this approach to richer
formulations, or as a simpler alternative to Approach~2.

\subsection{Limitations}
\label{sec:limitations}

Several known limitations qualify the conclusions of this study:

\begin{itemize}
    \item \textbf{Irreducible loss dominance at large scale.} At sufficiently
        large compute budgets, scaling properties are dominated entirely by the
        irreducible loss $E$. When token counts and model sizes at fixed compute
        budgets are large enough, the Chinchilla surface reaches $E$
        asymptotically and all training configurations become equally effective,
        meaning that extrapolations are irrelevant and compute-optimal training
        is no longer informed by scaling laws. We assume this study is only
        relevant to practitioners working in a regime where downstream model
        quality can still effectively be informed by scaling law extrapolations
        per the Chinchilla model.

    \item \textbf{Assumed correctness of the Chinchilla loss surface.} We
        assume the Chinchilla loss surface model
        $L(N, D) = E + A/N^\alpha + B/D^\beta$ is correct in practice. While
        there is substantial evidence legitimizing this
        model~\citep{chinchilla_robustness}, alternatives exist, including the
        Kaplan loss model~\citep{kaplan_scaling}, refined analytical surfaces
        like Farseer~\citep{farseer} and MuPT~\citep{mupt}, and
        agent-discovered functional forms~\citep{sld_agent}.

    \item \textbf{Qualitative characterization of published study errors.}
        Likely errors in published studies are characterized qualitatively
        rather than quantified. We believe the qualitative characterization is
        compelling enough on its own to justify that real IsoFLOP sampling
        pathologies occur in practice, but they are difficult to quantify
        precisely because they do not follow the convenient theoretical model we
        use for those pathologies in our simulations.
\end{itemize}

\appendix
\section*{Appendix}
\addcontentsline{toc}{section}{Appendix}
\renewcommand{\thefigure}{A\arabic{figure}}
\setcounter{figure}{0}
\renewcommand{\thetable}{A\arabic{table}}
\setcounter{table}{0}

\section{VPNLS Implementation Validation}
\label{sec:appendix_vpnls_validation}

We validated our VPNLS implementation against
\texttt{ml-scalefit}~\citep{optimal_data_mixtures}.
Both were run on the Chinchilla dataset (217 training points
with $C < 10^{21}$, inputs normalized by $N/10^6$ and $D/10^9$).
Table~\ref{tab:vpnls_validation} shows the fitted parameters.

Both Approach~3 variants use log-sum-exp (LSE) reparameterization to ensure
non-negativity of the additive loss components. They differ only in whether the
objective is evaluated on raw loss values (linear) or their logarithm
(log-scaled), as specified in the original Chinchilla formulation.

The \texttt{ml-scalefit} MSE configuration matches VPNLS to three decimal
places ($1 \times 10^{-3}$). Approach~3 with LSE
reparameterization but without log-scaled loss also produces identical results.
Only when log-scaled loss is used do the estimates diverge. This highlights
that both VPNLS and \texttt{ml-scalefit} (with MSE) are maximum likelihood
estimators under additive Gaussian error on the raw loss, not a multiplicative
error model as implied by the original Chinchilla Approach~3 log-loss
formulation. The Huber variant down-weights outliers and produces modestly
different estimates.

\begin{table}[H]
    \centering
    \small
    \begin{tabular}{lccccccc}
        \toprule
        Method & $E$ & $A$ & $B$ & $\alpha$ & $\beta$ & $a$ & $b$ \\
        \midrule
        VPNLS & 1.9051 & 4.0005 & 1.0509 & 0.3511 & 0.4587 & 0.5665 & 0.4335 \\
        Approach~3 (log-scaled loss) & 1.8926 & 4.0146 & 1.0571 & 0.3522 & 0.4430 & 0.5571 & 0.4429 \\
        Approach~3 (linear loss) & 1.9051 & 4.0005 & 1.0509 & 0.3511 & 0.4587 & 0.5665 & 0.4335 \\
        \midrule
        \texttt{ml-scalefit} (MSE) & 1.9051 & 4.0001 & 1.0509 & 0.3510 & 0.4588 & 0.5665 & 0.4335 \\
        \texttt{ml-scalefit} (Huber) & 1.8267 & 3.7066 & 1.0386 & 0.3240 & 0.4080 & 0.5573 & 0.4427 \\
        \bottomrule
    \end{tabular}
    \caption{VPNLS validation against \texttt{ml-scalefit} on the Chinchilla
    dataset. VPNLS and \texttt{ml-scalefit} (MSE) agree to three decimal places ($1 \times 10^{-3}$).
    Approach~3 with linear loss is identical to VPNLS; only the log-scaled loss
    variant diverges.}
    \label{tab:vpnls_validation}
\end{table}

\section{IsoFLOP Quality Control Pipeline}
\label{sec:appendix_isoflop_qc}

The IsoFLOP quality control pipeline applies these steps in this order:

\begin{itemize}
    \item \textbf{Pre-QC} removes duplicate and near-duplicate parameter values
    within each budget. Exact duplicates are disambiguated by keeping the point
    whose implied compute ($6ND$) is closest to the nominal budget (ties broken
    by minimum loss). Near-duplicates are identified by greedy binning in
    log-parameter space (tolerance 0.01) and disambiguated the same way. Budgets
    with fewer than 6 remaining points are then removed entirely.

    \item \textbf{Off Center} removes points that fall outside a symmetric
    window around each IsoFLOP curve's parabola minimum ($2.5\times$ the
    distance from the minimum to the nearer edge), targeting the off-center
    sampling bias identified in the main text. An Akima spline step also removes individual statistical outliers at
    this stage by fitting a leave-one-out spline to non-outlying points at
    each budget and flagging those with extreme MAD-based residuals. Across
    all experiments, every spline-identified point (28 of 28) is terminal,
    having no non-outlying point on at least one side in sorted parameter
    order. In practice, the spline step acts as a more aggressive complement
    to the conservative off-center filter.

    \item \textbf{Weak Curvature} first flags entire budgets with negative
    parabola curvature (quadratic coefficient $\leq 0$). For remaining budgets,
    it checks whether the curvature is significantly positive by testing whether
    the 95\% confidence interval on the quadratic coefficient includes zero.
    Budgets that fail either check are removed.

    \item \textbf{Post-QC} removes budgets that fell below the 6-point minimum
    after earlier filtering.
\end{itemize}

\begin{figure}[H]
    \centering
    \includegraphics[width=\textwidth]{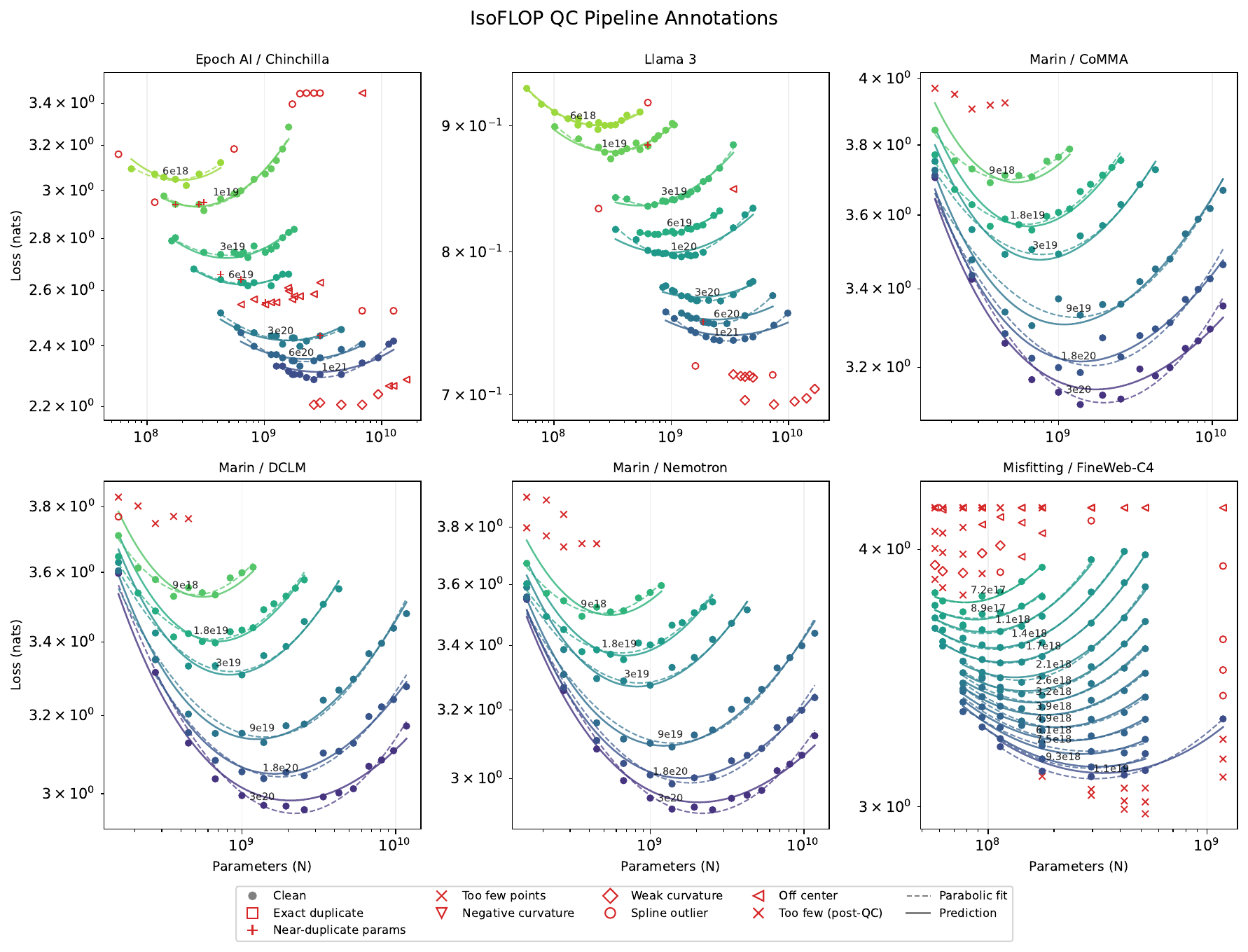}
    \caption{IsoFLOP curves for six experiments spanning different datasets
    and model families. Clean points (circles) and outlier points (markers
    by reason) are shown alongside parabolic fits (dashed) and Approach~3
    predictions with per-budget FLOP factor adjustment (solid). Experiments
    shown: Epoch~AI/Chinchilla, Llama~3, Marin/CoMMA, Marin/DCLM,
    Marin/Nemotron, and Misfitting/FineWeb-C4.}
    \label{fig:appendix_isoflop_qc}
\end{figure}

\section{IsoFLOP Samples with Noise}
\label{sec:appendix_noisy_isoflop}

\begin{figure}[H]
    \centering
    \includegraphics[width=\textwidth]{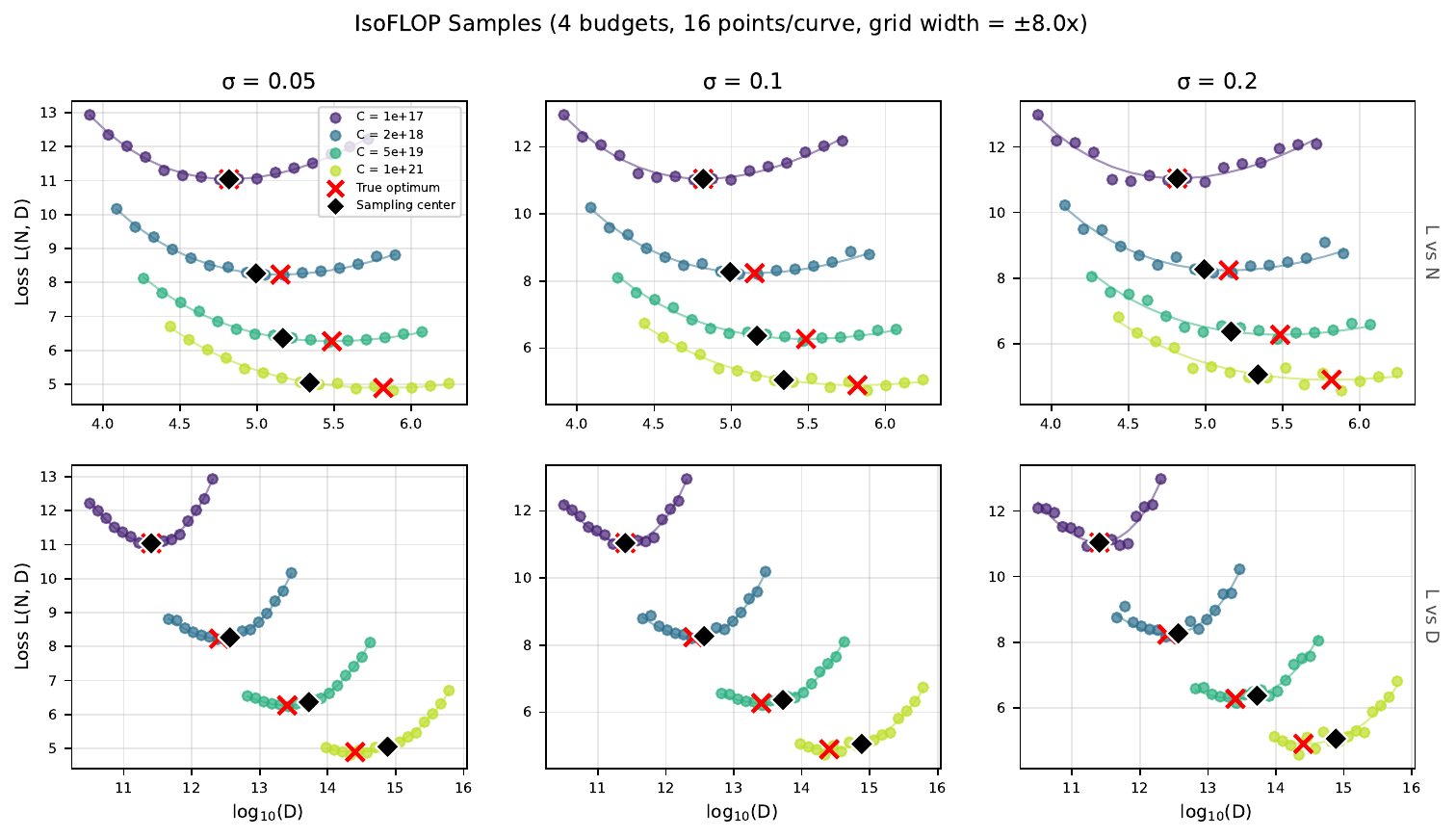}
    \caption{Noisy IsoFLOP samples used in the exponent inference comparison
    (Figure~\ref{fig:exponent_inference}). Columns correspond to noise levels
    ($\sigma = 0.05, 0.1, 0.2$); rows show loss versus $\log_{10}(N)$ (top)
    and $\log_{10}(D)$ (bottom). Scatter points are noisy observations; solid
    curves show the noiseless reference surface. Red $\times$ marks the true
    compute-optimal point at each budget; black diamonds mark the sampling
    centers, which drift away from the true optima at higher compute budgets.
    Surface: Asymmetric ($\alpha = 0.465$, $\beta = 0.155$), 4 budgets
    ($10^{17}$--$10^{21}$~FLOPs), 32 points per curve, $\pm 8\times$ grid.}
    \label{fig:appendix_noisy_isoflop}
\end{figure}

\section{Detailed Method Comparison}
\label{sec:appendix_method_comparison}

Full per-parameter, per-surface, per-sampling-range error breakdown for all
six method configurations.

\begin{figure}[H]
    \centering
    \includegraphics[width=\textwidth]{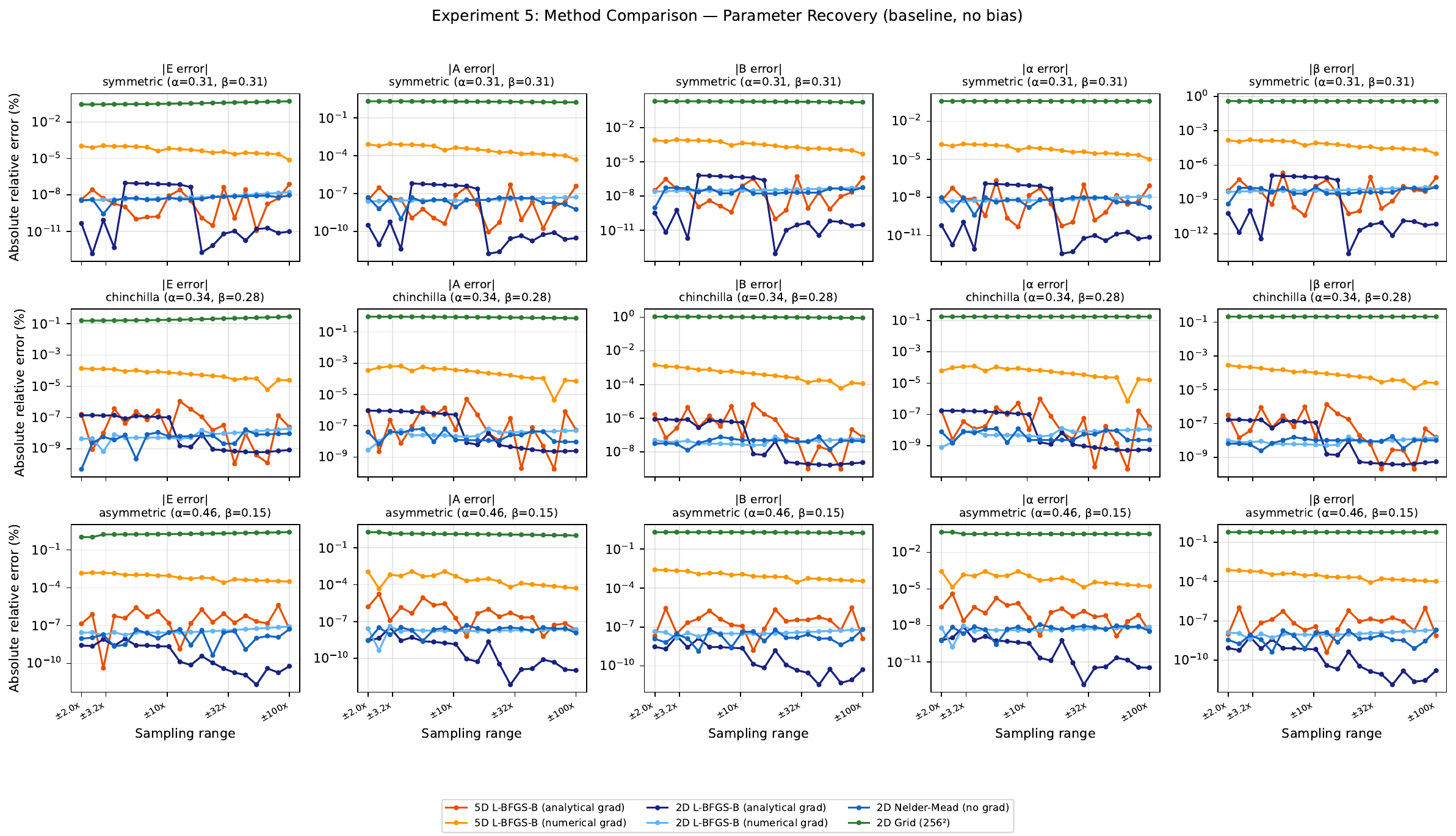}
    \caption{Per-parameter recovery error for six fitting methods across three
    loss surfaces and 20 sampling ranges (baseline, no bias). Each panel shows
    absolute relative error (\%) on a log scale versus sampling range, with one
    curve per method. Rows correspond to loss surfaces (symmetric, Chinchilla,
    high imbalance); columns correspond to parameters ($E$, $A$, $B$, $\alpha$,
    $\beta$).}
    \label{fig:appendix_method_detail}
\end{figure}

\section{Combined Extrapolation Error by Compute Budget}
\label{sec:appendix_combined_extrapolation}

Detailed view of $D^*$ extrapolation error as a function of compute budget,
showing how errors evolve from $10^{22}$ to $10^{25}$ FLOPs across sampling
ranges, loss surfaces, and bias configurations.

\begin{figure}[H]
    \centering
    \includegraphics[width=\textwidth]{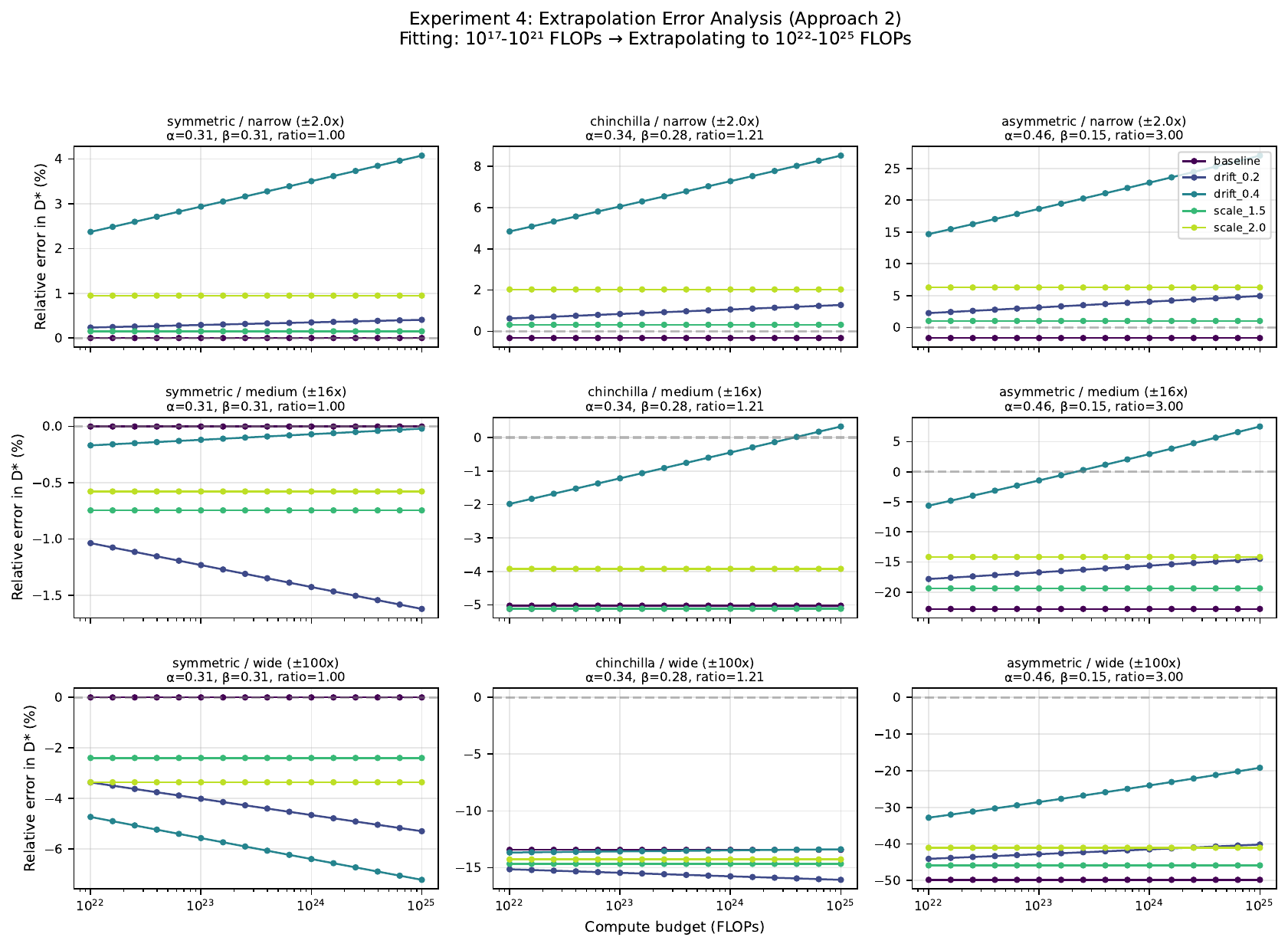}
    \caption{Relative error in compute-optimal token count $D^*$ when
    extrapolating from the fitting range ($10^{17}$--$10^{21}$ FLOPs) to higher
    compute budgets ($10^{22}$--$10^{25}$ FLOPs), with asymmetry and sampling
    biases acting simultaneously. Columns correspond to loss surfaces (symmetric,
    Chinchilla, Asymmetric); rows correspond to sampling ranges (narrow $\pm
    2\times$, medium $\pm 16\times$, wide $\pm 100\times$). The wide row uses an
    extreme grid width not employed in the main text, included here to further
    illustrate how far results can deviate with a misconfigured experiment. Each
    curve represents a different sampling bias configuration: baseline (no bias),
    two linear drift rates (drift\_0.2 and drift\_0.4, where the value is the
    $\log_{10}$ offset at the highest compute budget), and two constant center
    offsets (scale\_1.5 and scale\_2.0, where the value is the multiplicative
    factor applied at every budget). On symmetric surfaces, errors are driven
    entirely by off-center sampling; on asymmetric surfaces, the inherent surface
    bias adds a constant offset visible as the non-zero baseline curve.
    Drift-based biases produce errors that grow with extrapolation distance
    (steeper curves), while constant offsets and surface asymmetry produce flat
    errors.}
    \label{fig:appendix_combined_extrapolation}
\end{figure}

\section{Exponent Inference Error Breakdown}
\label{sec:appendix_exponent_errors}

\begin{figure}[H]
    \centering
    \includegraphics[width=\textwidth]{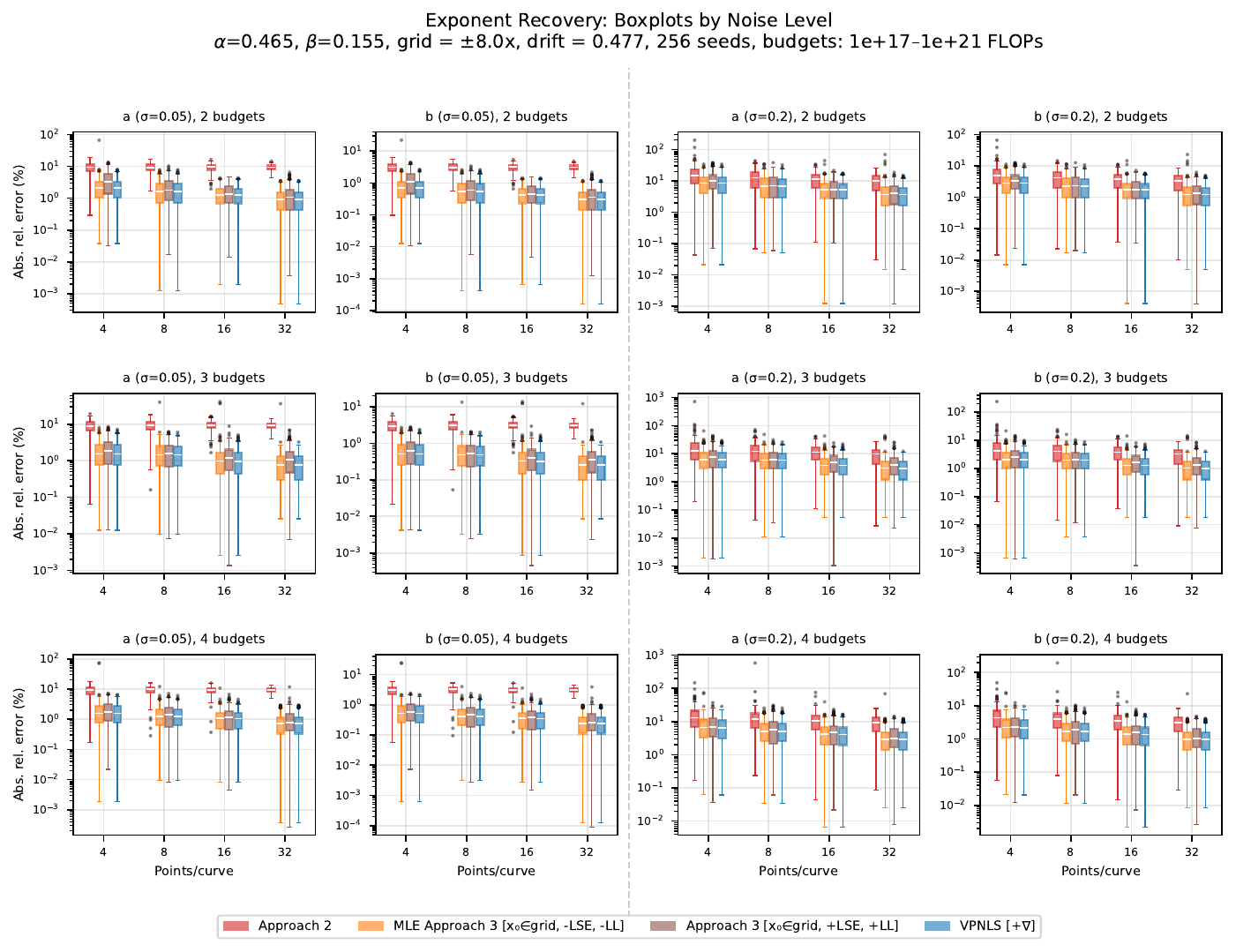}
    \caption{Per-condition breakdown of exponent inference errors from
    Figure~\ref{fig:exponent_inference}, showing 4 of 5 methods (excludes
    Naive Approach~3, which serves only as a negative control in the main
    figure). Rows correspond to number of compute budgets (2, 3, 4); left
    columns show the lowest noise level ($\sigma = 0.05$) and right columns
    show the highest ($\sigma = 0.2$), each split by exponent ($a$ and $b$).
    Within each panel, boxplots show absolute relative error (\%) on a log
    scale for each method at each points-per-curve setting (4, 8, 16, 32),
    over 256 noise realizations.}
    \label{fig:appendix_exponent_errors}
\end{figure}

\section{Data Efficiency Error Breakdown}
\label{sec:appendix_data_efficiency}

\begin{figure}[H]
    \centering
    \includegraphics[width=\textwidth]{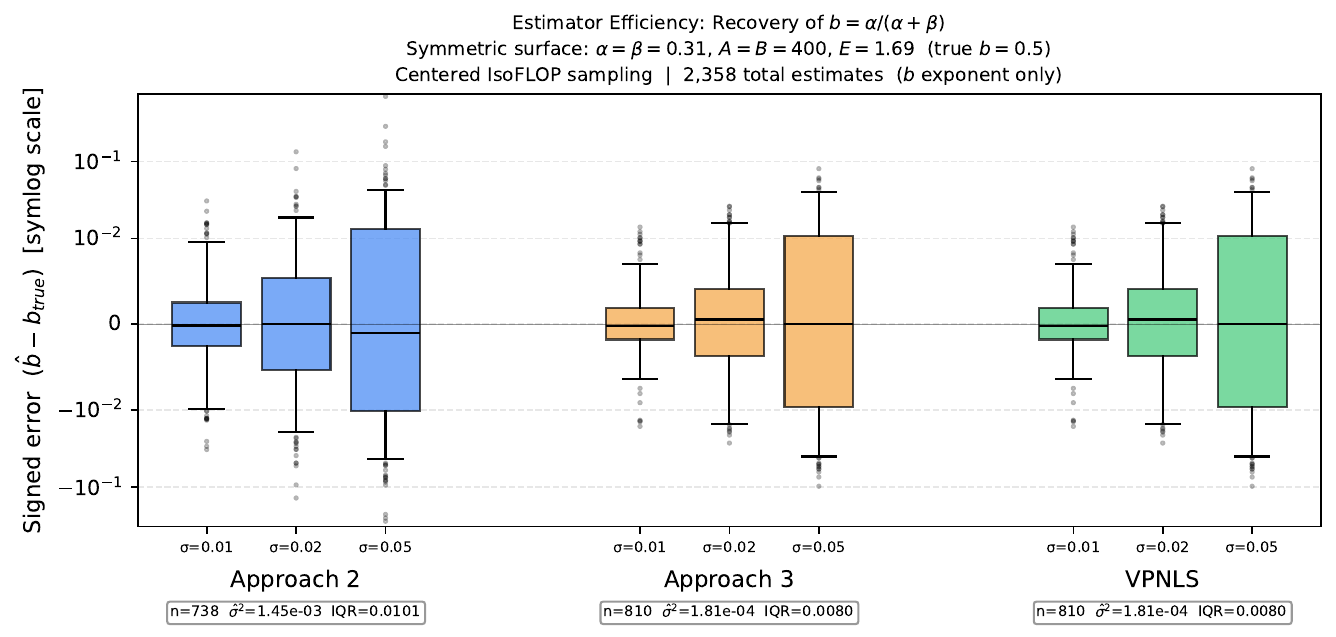}
    \caption{Per-noise-level breakdown of signed exponent errors from
    Figure~\ref{fig:data_efficiency}. Boxplots show the distribution of
    $\hat{b} - b_{\mathrm{true}}$ on a symlog scale for each method at each
    noise level. Pooled summary statistics (sample size, variance, IQR) are
    shown below each method group.}
    \label{fig:appendix_data_efficiency}
\end{figure}

\section{Published Scaling Exponents}
\label{sec:appendix_published_exponents}

Table~\ref{tab:published_exponents} collects published compute-allocation
exponents $a$ and $b$ (where $N^* \propto C^a$ and $D^* \propto C^b$) from
scaling law studies across a range of domains, architectures, and fitting
methods.

\begin{table}[H]
    \centering
    \small
    \begin{tabular}{llllccc}
        \toprule
        Source & Domain & Config & Method & $a$ & $b$ & $\beta/\alpha$ \\
        \midrule
        \multirow{5}{*}{DeepSeek~\citep{deepseek}}
            & \multirow{5}{*}{Web text}
            & OpenWebText2\textsuperscript{\textdaggerdbl} & \multirow{5}{*}{Appr.~2}
            & 0.73 & 0.27 & 2.70 \\
        & & MassiveText\textsuperscript{\textdaggerdbl} & & 0.49 & 0.51 & 0.96 \\
        & & DeepSeek (early) & & 0.450 & 0.550 & 0.82 \\
        & & DeepSeek (current) & & 0.524 & 0.476 & 1.10 \\
        & & DeepSeek (OWT2) & & 0.578 & 0.422 & 1.37 \\
        \midrule
        EHR~\citep{ehr_scaling} & Health records & Transformer & Appr.~2
            & 0.58 & 0.44 & 1.32 \\
        \midrule
        \multirow{4}{*}{Evo~\citep{evo}\textsuperscript{\textdagger}}
            & \multirow{4}{*}{DNA sequences}
            & Transformer++ & \multirow{4}{*}{Appr.~2}
            & 0.552 & 0.551 & 1.00 \\
        & & Mamba & & 0.388 & 0.487 & 0.80 \\
        & & Hyena & & 0.504 & 0.499 & 1.01 \\
        & & StripedHyena & & 0.483 & 0.539 & 0.90 \\
        \midrule
        \multirow{3}{*}{IL~\citep{il_scaling}}
            & \multirow{3}{*}{Atari (RL)}
            & NetHack & \multirow{3}{*}{Appr.~2}
            & 0.61 & 0.39 & 1.56 \\
        & & Battle Zone & & 0.58 & 0.51 & 1.14 \\
        & & Breakout & & 0.74 & 0.46 & 1.61 \\
        \midrule
        DiT~\citep{dit_scaling} & Image-text & Diffusion Trans. & Appr.~2
            & 0.568 & 0.432 & 1.32 \\
        \midrule
        DLM~\citep{dlm_scaling} & Web text & Discrete diffusion & Appr.~2
            & 0.589 & 0.411 & 1.43 \\
        \midrule
        Audio~\citep{audio_scaling} & Audio + text & Transformer & Appr.~2
            & 0.367 & 0.579 & 0.63 \\
        \midrule
        \multirow{2}{*}{BLM~\citep{beyond_language_modeling}}
            & \multirow{2}{*}{Text, video, image}
            & Language & \multirow{2}{*}{Appr.~2}
            & 0.47 & 0.53 & 0.89 \\
        & & Vision & & 0.37 & 0.63 & 0.59 \\
        \midrule
        \multirow{2}{*}{Protein PLM~\citep{protein_plm_scaling}}
            & \multirow{2}{*}{Protein seq.}
            & CLM & \multirow{2}{*}{Appr.~2}
            & 0.578 & 0.422 & 1.37 \\
        & & MLM & & 0.776 & 0.230 & 3.37 \\
        \midrule
        \multirow{2}{*}{xLSTM~\citep{xlstm_scaling}}
            & \multirow{2}{*}{Web text}
            & Transformer & \multirow{2}{*}{Appr.~2}
            & 0.575 & 0.424 & 1.36 \\
        & & xLSTM & & 0.547 & 0.417 & 1.31 \\
        \midrule
        \multirow{3}{*}{NMM~\citep{nmm_scaling}}
            & \multirow{3}{*}{Multimodal}
            & Early fusion (avg.) & \multirow{3}{*}{Appr.~3}
            & 0.526 & 0.473 & 1.11 \\
        & & Late fusion & & 0.636 & 0.462 & 1.38 \\
        & & Sparse early fusion & & 0.361 & 0.656 & 0.55 \\
        \midrule
        \multirow{3}{*}{OLMo~\citep{olmo_hybrid}}
            & \multirow{3}{*}{Web text}
            & Transformer & \multirow{3}{*}{Appr.~3}
            & 0.458 & 0.542 & 0.85 \\
        & & Hybrid & & 0.492 & 0.508 & 0.97 \\
        & & Pure GDN & & 0.554 & 0.446 & 1.24 \\
        \midrule
        ProGen3~\citep{progen3} & Protein seq. & ProGen3 & Kaplan
            & & & 1.48\textsuperscript{*} \\
        \midrule
        Dense Retrieval~\citep{dense_retrieval_scaling} & Query-doc pairs & Transformer & Kaplan
            & & & 2.34\textsuperscript{*} \\
        \bottomrule
    \end{tabular}
    \vspace{4pt}

    {\footnotesize
    \textsuperscript{*} Reported directly. The Kaplan method estimates
    $\alpha$ and $\beta$ separately, so $\beta/\alpha$ is obtained without
    assuming the Chinchilla functional form for $a$ and $b$.\\
    \textsuperscript{\textdagger} Exponents extracted from Figure S5
    data.\footnote{\url{https://github.com/eric-czech/evo-scaling-law-extraction}}\\
    \textsuperscript{\textdaggerdbl} Exponents originally from
    Kaplan et al.~\citep{kaplan_scaling} (OpenWebText2) and
    Hoffmann et al.~\citep{chinchilla} (MassiveText), as reported in DeepSeek.
    }
    \caption{Published compute-allocation exponents from scaling law studies
    across domains. Columns $a$ and $b$ are the allocation exponents governing
    optimal model size ($N^* \propto C^a$) and dataset size
    ($D^* \propto C^b$). The $\beta/\alpha$ column gives the ratio of
    loss-surface exponents, computed as $a/b$ for Approach~2 and~3 estimates
    or reported directly for Kaplan estimates.}
    \label{tab:published_exponents}
\end{table}

\section{Progressive Filtering for Chinchilla}
\label{sec:appendix_progressive_chinchilla}

\begin{figure}[H]
    \centering
    \includegraphics[width=\textwidth]{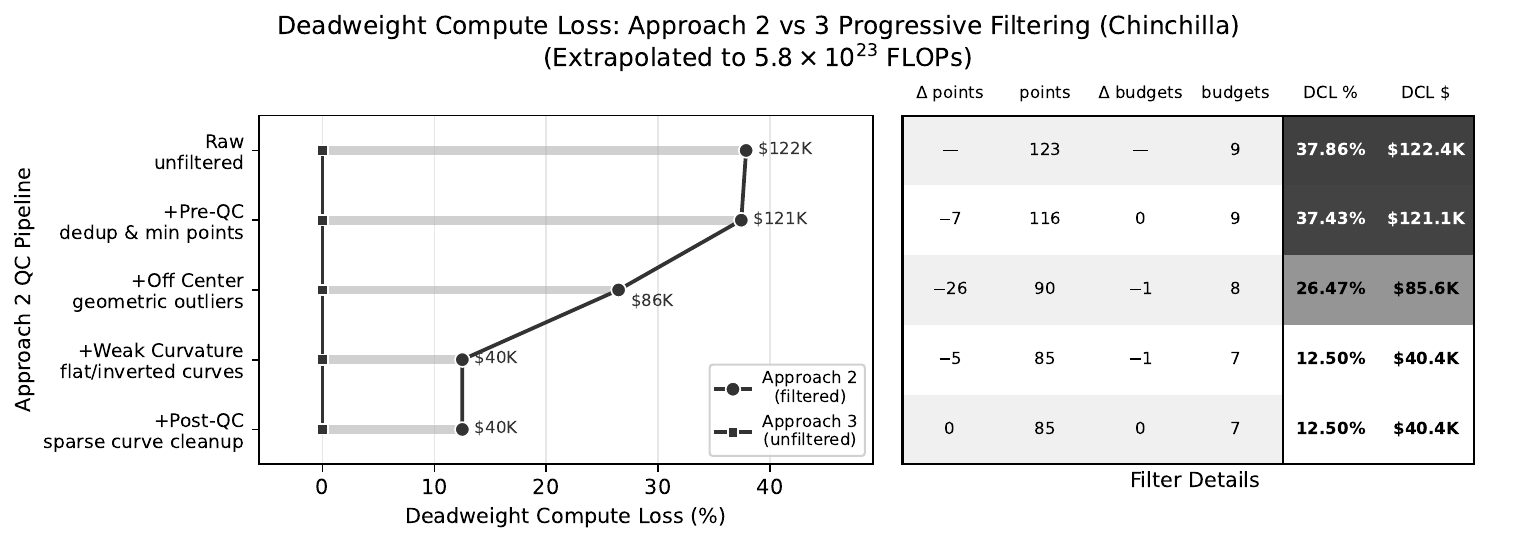}
    \caption{Progressive quality control filtering on Chinchilla data,
    analogous to Figure~\ref{fig:progressive-filter} for Llama~3. DCL is
    measured against an Approach~3 surface fit to unfiltered data and
    extrapolated to $5.8\times10^{23}$~FLOPs. Starting from 37.9\% DCL on
    raw data, the off-center and weak-curvature filters reduce DCL to 12.5\%.}
    \label{fig:appendix_progressive_chinchilla}
\end{figure}

\section{Residual Distributions by Budget}
\label{sec:appendix_residual_distributions}

\begin{figure}[H]
    \centering
    \includegraphics[width=\textwidth]{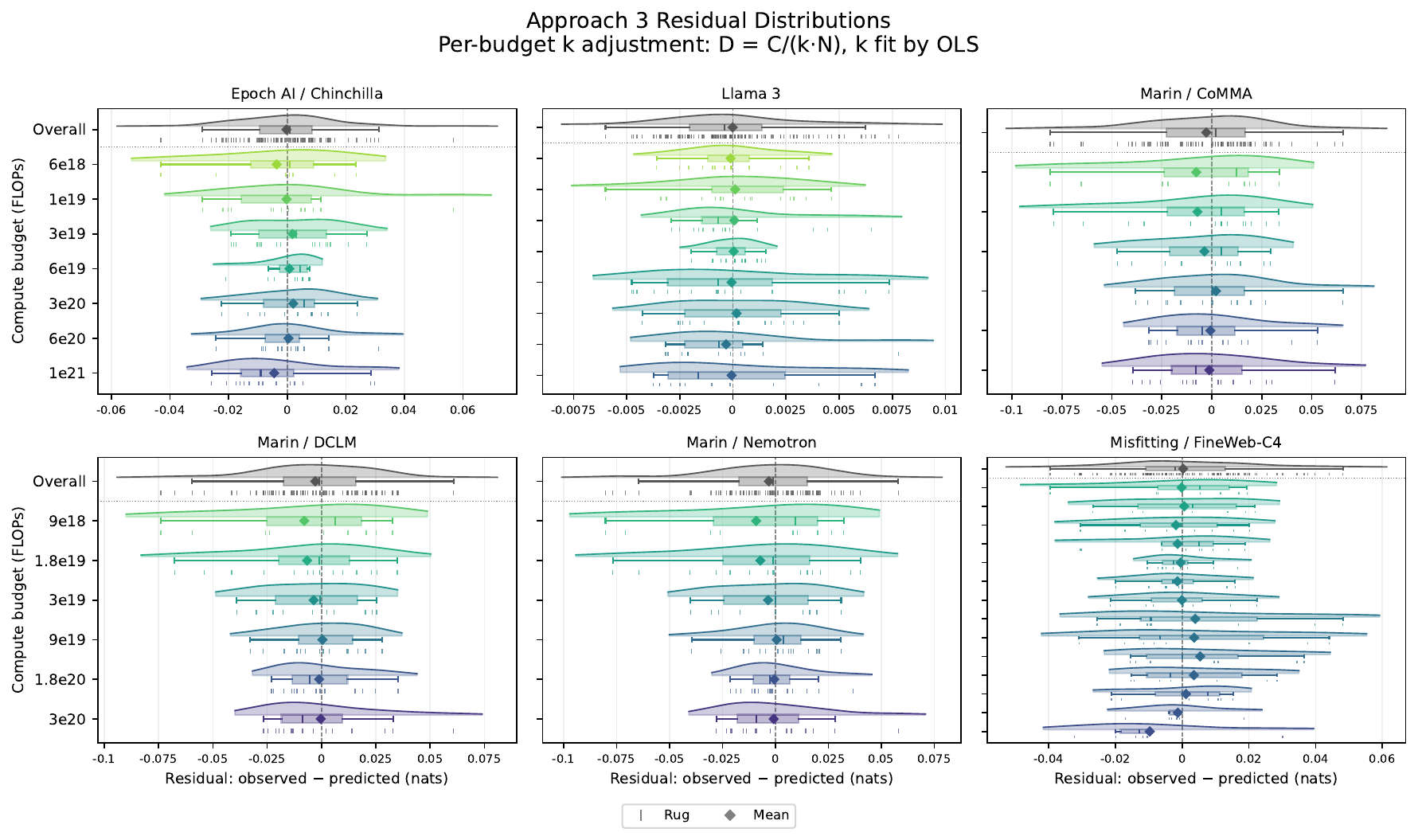}
    \caption{Per-budget residual distributions after Approach~3 fitting with
    per-budget FLOP factor adjustment, across six experiments. Each row shows
    the residuals for one compute budget as a layered rug plot, boxplot, and
    KDE, with an overall row pooling all clean points. The zero-residual
    reference line (dashed) indicates perfect prediction.}
    \label{fig:appendix_residual_distributions}
\end{figure}

\section{Residual Variance Summary}
\label{sec:appendix_residual_variance}

\begin{figure}[H]
    \centering
    \includegraphics[width=\textwidth]{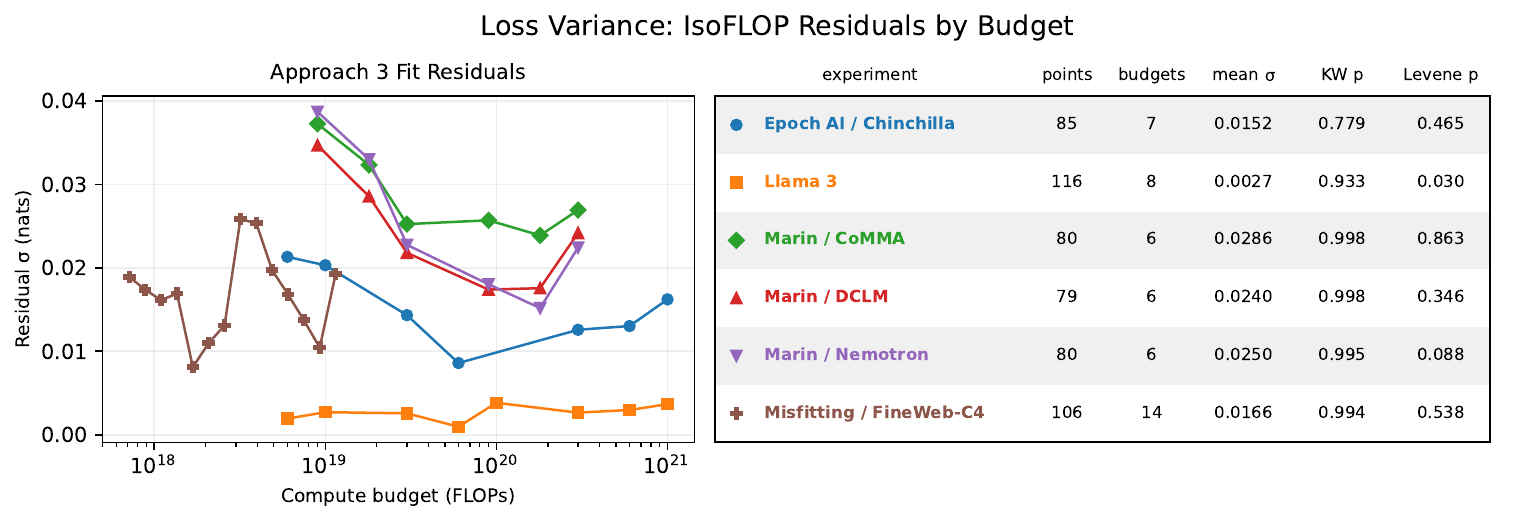}
    \caption{Per-budget residual standard deviation (left) and summary
    statistics (right) for six experiments. Kruskal-Wallis tests whether
    residual locations differ across budgets. Levene tests whether residual
    variances differ across budgets. High p-values in most experiments
    indicate that a budget-independent noise model is a reasonable
    approximation for simulation purposes.}
    \label{fig:appendix_residual_variance}
\end{figure}

\section{Contents}
\label{sec:appendix_contents}
\tableofcontents

\end{document}